%% file: main.tex
\documentclass[final,12pt]{clear2024}
\PassOptionsToPackage{numbers, compress}{natbib}

\usepackage[utf8]{inputenc} %
\usepackage[T1]{fontenc}    %
\usepackage{hyperref}       %
\usepackage{url}            %
\usepackage{booktabs}       %
\usepackage{amsfonts}       %
\usepackage{nicefrac}       %
\usepackage{microtype}      %
\usepackage{xcolor}         %
\usepackage{wrapfig}

\usepackage{amsmath}
\usepackage{amsfonts}

\usepackage{natbib}
\usepackage{tikz}
\usepackage{booktabs}
\usepackage{graphics}
\usepackage{mathabx}
\usepackage{algorithm}
\usepackage{algorithmic}
\usepackage{clrscode3e}

\usepackage{inconsolata}
\usepackage{mathptmx}
\usepackage{wasysym}

\newcommand{\secref}[1]{Section~\ref{#1}}
\newcommand{\appref}[1]{Appendix~\ref{#1}}

\newcommand{\Defref}[1]{Definition~\ref{#1}}
\newcommand{\defref}[1]{Definition~\ref{#1}}
\newcommand{\Figref}[1]{Figure~\ref{#1}}

\newcommand{\Tabref}[1]{Table~\ref{#1}}

\newcommand{\ReLU}{\mathsf{ReLU}}
\newcommand{\softmax}{\mathbf{softmax}}
\newcommand{\methodabbr}{DAS}
\newcommand{\intaccmetricabbr}{IIA}
\newcommand{\methodname}{distributed alignment search}

\newcommand{\intinv}{\textsc{II}}
\newcommand{\dintinv}{\textsc{DII}}

\usepackage[]{titlesec}
\titlespacing*{\section}{0pt}{3pt}{3pt}
\titlespacing*{\subsection}{0pt}{3pt}{3pt}

\usetikzlibrary{shapes, arrows, calc,matrix,positioning}

\title[Finding Distributed Alignments]{Finding Alignments Between Interpretable\\ Causal Variables and Distributed Neural Representations}

\clearauthor{%
  Atticus Geiger$^{\ast}$$^{\diamondsuit}$, %
  Zhengxuan Wu\thanks{Equal contribution.}, %
  \textbf{Christopher Potts},  %
  \textbf{Thomas Icard}, and %
  \textbf{Noah D.~Goodman} %
  \\[1ex] 
  Pr(Ai)$^2$R Group$^{\diamondsuit}$ Stanford University \\
  \texttt{\{atticusg, wuzhengx, cgpotts, icard, ngoodman\}@stanford.edu} 
}

\begin{document}

\maketitle

\begin{abstract}
Causal abstraction is a promising theoretical framework for explainable artificial intelligence that defines when an interpretable high-level causal model is a faithful simplification of a low-level deep learning system. However, existing causal abstraction methods have two major limitations: they require a brute-force search over alignments between the high-level model and the low-level one, and they presuppose that variables in the high-level model will align with disjoint sets of neurons in the low-level one. In this paper, we present \textit{\methodname}\ (\methodabbr), which overcomes these limitations. In \methodabbr, we find the alignment between high-level and low-level models using gradient descent rather than conducting a brute-force search, and we allow individual neurons to play multiple distinct roles by analyzing representations in non-standard bases---\textit{distributed} representations. Our experiments show that \methodabbr\ can discover internal structure that prior approaches miss. Overall, \methodabbr\ removes previous obstacles to uncovering conceptual structure in trained neural nets.
\end{abstract}

\section{Introduction}\label{sec:intro}

Can an interpretable symbolic algorithm be used to faithfully explain a complex neural network model? This is a key question for interpretability; a positive answer can provide guarantees about how the model will behave, and a negative answer could lead to fundamental concerns about whether the model will be safe and trustworthy.  

Causal abstraction provides a mathematical framework for precisely characterizing what it means for any complex causal system (e.g., a deep learning model) to implement a simpler causal system (e.g., a symbolic algorithm) \citep{Rubinstein2017, beckers20a, Massida}.  For modern AI models, the fundamental operation for assessing whether this relationship holds in practice has been the \textit{interchange intervention} (also known as activation patching), in which a neural network is provided a `base' input, and sets of neurons are forced to take on the values they would have if different `source' inputs were processed \citep{Geiger-etal-2020, vig2020causal, finlayson-2021-causal, meng2022locating}. The counterfactuals that these interventions create are the basis for causal inferences about model behavior.

\citet{Geiger:2021} show that the relevant causal abstraction relation obtains when interchange interventions on aligned high-level variables and low-level variables have equivalent effects. This ideal relationship rarely obtains in practice, but the proportion of interchange interventions with the same effect (\textit{interchange intervention accuracy}; \intaccmetricabbr) provides a graded notion, and \citet{geigericard}  formally ground this metric in the theory of approximate causal abstraction. \citeauthor{geigericard} also use causal abstraction theory as a unified framework for a wide range of recent intervention-based analysis methods \citep{vig2020causal,Csordas2021,CausalLM,Ravfogel:2020, Elazar-etal-2020,de2021sparse,abraham-etal-2022-cebab,olah2020zoom,olsson2022context,chan2022causal}.

Causal abstraction techniques have been applied to diverse problems \citep{Geiger-etal:2019,Geiger-etal-2020,Li2021,Huang-etal:2022}. However, previous applications have faced two central challenges. First, causal abstraction requires a computationally intensive brute-force search process to find optimal alignments between the variables in the high-level model and the states of the low-level one. Where exhaustive search is intractable, we risk missing the best alignment entirely. 
Second, these prior methods are \textit{localist}: they artificially limit the space of possible alignments by presupposing that high-level causal variables will be aligned with disjoint groups of neurons. There is no reason to assume this a priori, and indeed much recent work in model explanation (see especially \citealt{Ravfogel:2020,pmlr-v162-ravfogel22a, Elazar-etal-2020,olah2020zoom,olsson2022context}) is converging on the insight of \citet{Smolensky1986},  \citet{PDP1}, and \citet{PDP2} that individual neurons can play multiple conceptual roles. \citet{Smolensky1986} identified \textit{distributed neural representations} as ``patterns'' consisting of linear combinations of unit vectors.

In the current paper, we propose \methodname\ (\methodabbr), which overcomes the above limitations of prior causal abstraction work. In \methodabbr, we find the best alignment via \textit{gradient descent} rather than conducting a brute-force search. In addition, we use \textit{distributed interchange interventions}, which are ``soft'' interventions in which the causal mechanisms of a group of neurons are edited such that (1) their values are rotated with a change-of-basis matrix, (2) the targeted dimensions of the rotated neural representation are fixed to be the corresponding values in the rotated neural representation created for the source inputs, and (3) the representation is rotated back to the standard neuron-aligned basis. The key insight is that viewing a neural representation through an alternative basis that is not aligned with individual neurons can reveal interpretable dimensions \citep{Smolensky1986}.
 
In our experiments, we evaluate the capabilities of \methodabbr\ to provide faithful and interpretable explanations with two tasks that have obvious interpretable high-level algorithmic solutions with two intermediate variables. In both tasks, the distributed alignment learned by \methodabbr\ is as good or better than both the closest localist alignment and the best localist alignment in a brute-force search.

In our first set of experiments, we focus on a hierarchical equality task that has been used extensively in developmental and cognitive psychology as a test of relational reasoning \citep{Premack:1983,thompson:1997,Geiger:2022aa}: the inputs are sequences $[w, x, y, z]$, and the label is given by $(w=x) = (y=z)$. We train a simple feed-forward neural network on this task and show that it perfectly solves the task. Our key question: does this model implement a program that computes $w=x$ and $y=z$ as intermediate values, as we might hypothesize humans do? Using \methodabbr, we find a distributed alignment with 100\% \intaccmetricabbr. In other words, the network is perfectly abstracted by the high-level model; the distinction between the learned neural model and the symbolic algorithm is thus one of implementation.

Our second task models a natural language inference dataset \citep{Geiger-etal-2020} where the inputs are premise and hypothesis sentences $(p,h)$ that are identical but for the words $w_p$ and $w_h$; the label is either \textit{entails} ($p$ makes $h$ true) or \textit{contradicts}/\textit{neutral} ($p$ makes $h$ false). We fine-tune a pretrained language model to perfectly solve the task. With \methodabbr, we find a perfect alignment (100\% \intaccmetricabbr) to a causal model with a binary variable for the entailment relation between the words $w_p$ and $w_h$ (e.g., \textit{dog} entails \textit{mammal}). 

In both our sets of experiments, the \methodabbr\ analyses reveal perfect abstraction relations. However, we also identify an important difference between them. In the NLI case, the entailment relation can be decomposed into representations of $w_p$ and $w_h$.  What appears to be a representation of lexical entailment is, in this case, a ``data structure'' containing two representations of word identity, rather than an encoding of their entailment relation. By contrast, the hierarchical equality models learn representations of $w=x$ and $y=z$ that cannot be decomposed into representations of $w$, $x$, $y$ and $z$. In other words, these relations are entirely abstracted from the entities participating in the relation; \methodabbr\ reveals that the neural network truly implements a symbolic, tree-structured algorithm.

 \section{Related Work}
A theory of \emph{causal abstraction} specifies exactly when a `high-level causal model' can be seen as an abstract characterization of some `low-level causal model' \citep{Simon,Chalupka, Rubinstein2017, beckers20a}. The basic idea is that high-level variables are associated with (potentially overlapping) sets of low-level variables that summarize their causal mechanisms with respect to a set of hard or soft interventions \citep{Massida}. In practice, a graded notion of \emph{approximate} causal abstraction is often more useful \citep{beckers20a, RischelWeichwald, geigericard}.

 \citet{geigericard} argue that causal abstraction is a generic theoretical framework for providing \textit{faithful} \citep{Jacovi2020, Lyu} and \textit{interpretable} \citep{Lipton18} explanations of AI models and show that LIME \citep{Ribeiro2016}, causal effect estimation \citep{abraham-etal-2022-cebab, CausalLM}, causal mediation analysis \citep{vig2020causal,Csordas2021,de2021sparse}, iterated nullspace projection \citep{Ravfogel:2020, Elazar-etal-2020}, and circuit-based explanations \citep{olah2020zoom,olsson2022context,wang2022interpretability, chan2022causal} can all be understood as causal abstraction analysis. 

Interchange intervention training (IIT) objectives are minimized when a high-level causal model is an abstraction of a neural network under a given alignment
\citep{geiger-etal-2021-iit, wu-etal-2022-cpm, Huang-etal:2022}. In this paper, we use IIT objectives to learn an alignment between a high-level causal model and a deep learning model.

\section{Methods}\label{sec:methods}

\input{methods-alt.tex}

\subsection{General Experimental Setup}

We illustrate the value of \methodabbr\ by analyzing feed-forward networks trained on a hierarchical equality and pretrained Transformer-based language models \citep{Vaswani-etal:2017} fine-tuned on a natural language inference task. Our evaluation paradigm is as follows:
\begin{enumerate}\setlength{\itemsep}{0pt}
\item Train the neural network $\mathcal{N}$ to solve the task. In all experiments, the neural models achieve perfect accuracy on both training and testing data. 
\item Create interchange intervention training datasets using a high-level causal model. Each example consists of a base input, one or more source inputs, high-level causal variables targetted for intervention, and a counterfactual gold label that will be output by the network if the interchange intervention has the hypothesized effect on model behavior. This gold label is a counterfactual output of the high-level model we will align with the network. (See \appref{app:training-data-construction} for details)
\item Optimize an orthogonal matrix to learn a distributed alignment for each high-level model that maximizes \intaccmetricabbr\ using the training objective in Def.~\ref{def:trainobj}. We experiment with different hidden dimension sizes for our low-level model and different intervention site sizes (dimensionality of low-level subspaces) and locations (the layer where the intervention happens). (See \appref{app:reproducibility} for details)
\item Evaluate a baseline that brute-force searches through a discrete space of alignments and selects the alignment with the highest \intaccmetricabbr. We search the space of alignments by aligning each high-level variable with groups of neurons in disjoint sliding windows. (See \appref{app:greedy-search} for details)
\item Evaluate the localist alignment ``closest'' to the learned distributed alignment. The rotation matrix for the localist alignment will be axis-aligned with the standard basis, possibly permuting and reflecting unit axes. (See \appref{app:localist} for details)
\item Determine whether each distributed representation aligned with high-level variables can be decomposed into multiple representations that encode the identity of the input values to the variable's causal mechanism. We do this by learning a second rotation matrix that decomposes learned distributed representation, holding the first rotation matrix fixed. (See \appref{app:subspace-search} for details)
\end{enumerate} 

The codebase used to run these experiments is at\footnote{\url{https://github.com/atticusg/InterchangeInterventions/tree/zen}}. We have replicated the hierarchical equality experiment using the Pyvene library at\footnote{\url{https://github.com/stanfordnlp/pyvene/blob/main/tutorials/advanced_tutorials/DAS_Main_Introduction.ipynb}}.

\section{Hierarchical Equality Experiment}\label{sec:equality}

We now illustrate the power of \methodabbr\ for analyzing networks designed to solve a hierarchical equality task. We concentrate on analyzing a trained feed-forward network. 

A \emph{basic} equality task is to determine whether a pair of objects are the same ($x = y$). A \emph{hierarchical} equality task is to determine whether a pair of pairs of objects have identical relations: $(w=x) = (y=z)$. Specifically, the input to the task is two pairs of objects and the output is $\mathsf{True}$ if both pairs are equal or both pairs are unequal and $\mathsf{False}$ otherwise. For example, $(A,A,B,B)$ and $(A,B,C,D)$ are both assigned $\mathsf{True}$ while $(A,B,C,C)$ is assigned $\mathsf{False}$.

\subsection{Low-Level Neural Model}
We train a three-layer feed-forward network with ReLU activations to perform the hierarchical equality task. Each input object is represented by a randomly initialized vector. Specifically, our model has the following architecture where $k$ is the number of layers.
\[ h_1     = \ReLU([x_1;x_2;x_3;x_4]W_1 + b_{1})\; \; \; \; \; \; \; \; \; h_{j-1} = \ReLU(h_{j}W_{j} + b_{j})\; \; \;\; \; \; \; \; \; y= \softmax(h_{k}W_{k} + b_{k})\]
The input vectors are in $\mathbb{R}^n$, the biases are in $\mathbb{R}^{4n}$, and the weights are in $\mathbb{R}^{4n \times 4n}$. We evaluate our model on held-out random vectors unseen during training, as in \citealt{Geiger:2022aa}.

\input{tables/eq-results.tex}

\subsection{High-Level Models} 
We use \methodabbr\ to evaluate whether trained neural networks have achieved the natural solution to the hierarchical equality task where the left and right equality relations are computed and then used to predict the final label (\Figref{fig:higheqmod}).

\begin{wrapfigure}{r}{0.4\textwidth}
\centering
    \resizebox{0.25\textwidth}{!}{%
    \input{graphs/1234-large.tex} 
    }
    \caption{A causal model that computes the hierarchical equality task.}
 \label{fig:higheqmod}
\end{wrapfigure}

However, evaluating this high-level model alone is insufficient, as there are obviously many other high-level models of this task. To further contextualize our results, we also consider two alternatives: 
a high-level model where only the equality relation of the first pair is represented and a high-level model where the lone intermediate variable encodes the identity of the first input object (leaving all computation for the final step). These alternative high-level models also solve the task perfectly.

\subsection{Discussion} The \intaccmetricabbr\ results achieved by the best alignment for each high-level model can be seen in \Tabref{tab:equalityresults}. The best alignments found are with the `Both Equality Relations' model that is widely assumed in the cognitive science literature. For all causal models, \methodabbr\ learns a more faithful alignment (higher \intaccmetricabbr) than a brute-force search through localist alignments. This result is most pronounced for `Both Equality Relations', where \methodabbr\ learns perfect or near-perfect alignments under a number of settings, whereas the best brute-force alignment achieves only 0.60 and the best localist alignment achieves only 0.73.
Finally, the distributed representation of left equality could not be decomposed into a representation of the first argument identity. We see this in the very low performance of the `Identity Subspace of Left Equality' results. This indicates that models are truly learning to encode an abstract equality relation, rather than merely storing the identities of the inputs.

\subsection{Analyzing a Randomly Initialized Network}

\begin{wrapfigure}{r}{0.4\textwidth}
\centering
    \resizebox{1.0\linewidth}{!}{%
    \begin{tabular}{ccc}
    \multicolumn{3}{c}{Both Equality Relations} \\
    \multicolumn{3}{c}{\input{graphs/1234}}\\
    Hidden size & Intervention size & \textbf{Layer 1}\\
    \midrule
    $|\mathbf{N}| =16$ & $k = 8$  &  0.50  \\
    $|\mathbf{N}| =64$ & $k = 32$  &  0.50 \\
    $|\mathbf{N}| =256$ & $k = 128$  &  0.51 \\
    $|\mathbf{N}| =1028$ & $k = 512$  &  0.55 \\
    $|\mathbf{N}| =4096 $ & $k = 2048$  &  0.64 \\
    \bottomrule
    \end{tabular}}
    \label{tab:random}
    \caption{\methodabbr\ on a random network with a 16 dimension input. An oversized hidden dimension allows DAS to manipulate the model behavior by searching through a large space of random mechanisms.}
\end{wrapfigure}
To calibrate intuitions about our method, we evaluate the ability of \methodabbr\ to optimize for interchange intervention accuracy on a frozen randomly initialized networks that achieves chance accuracy (50\%) on the hierarchical equality task. This investigates the degree to which random causal structures can be used to systematically manipulate the counterfactual behavior of the network. 
We evaluate networks with different hidden representation sizes while holding the four input vectors fixed at $4$ dimensions, under the hypothesis that more hidden neurons create more random structure that \methodabbr\ can search through. These results are summarized in \Tabref{tab:random}. Observe that, in small networks, there is no ability to increase interchange intervention accuracy. However, as we increase the size of the hidden representation to be orders of magnitude larger than the input dimension of 16, the interchange intervention accuracy increases. This confirms our hypothesis and serves as a check that demonstrates \methodabbr\ cannot construct entirely new behaviors from random structure.

\begin{figure}[t]
\subfigure[Two MoNLI examples.]{
\resizebox{0.65\textwidth}{!}{
    \begin{tabular}{lc}
      \toprule
      \textbf{Sentence Pairs} & \textbf{Label} \\
      \midrule
      \emph{premise}: A man is talking to someone in a taxi. & \textit{entails} \\
      \emph{hypothesis}: A man is talking to someone in a car. & \\
      \midrule
      \emph{premise}: The people are \textbf{not} playing sitars. & \textit{neutral} \\
      \emph{hypothesis}: The people are \textbf{not}  playing instruments. & \\
      \bottomrule
    \end{tabular}
    }
    \label{fig:nliex} 
}
\hfill
\subfigure[A simple program that solves MoNLI.]{
\begin{minipage}{0.3\textwidth}
\footnotesize
$\textsc{MoNLI}(\textbf{p}, \textbf{h})$
\begin{codebox}
\li $\textit{lexrel} \leftarrow \proc{get-lexrel}(\textbf{p}, \textbf{h})$
\li $\textit{neg} \leftarrow \proc{contains-not}(\textbf{p}, \textbf{h})$
\li $\textbf{if}$ \textit{neg}:
\li \; \; \Return $\proc{reverse}(\textit{lexrel} )$ \End
\li \Return $\textit{lexrel}$
\end{codebox}
\label{fig:nliprog}
\end{minipage}
}
\caption{Monotonicity NLI task examples and high-level model.}
\end{figure}

\section{Monotonicity NLI Experiment}

In our second experiment, we analyze a BERT model fine-tuned on the Monotonicity Natural Language Inference (MoNLI) benchmark \citep{Geiger-etal-2020}. A MoNLI example consists of a premise sentence and hypothesis sentence and the output label is \textit{entails} when the premise makes the hypothesis true, and \textit{neutral} otherwise. Two examples are in~\Figref{fig:nliex}.
Every example is such that a single word $w_p$ in the premise sentence was changed to a hypernym (more general term) or hyponym (more specific term) $w_h$ to create the hypothesis. About half of MoNLI examples contain a negation that scopes over the word replacement site, and the remaining examples have no negation. When no negation is present, the label for a premise--hypothesis pair is the lexical relation. When negation is present, the label for a premise--hypothesis pair is the reverse of the lexical relation.

\subsection{Low-Level Neural Model} We fine-tune an uncased BERT-base model~\citep{devlin-etal-2019-bert} finetuned on the MultiNLI dataset \citep{N18-1101}.\footnote{The parameters are provided by the Hugging Face $\mathtt{transformers}$ library \citep{Wolf2019HuggingFacesTS}, downloaded from \url{https://huggingface.co/ishan/bert-base-uncased-mnli}.} Our BERT model has 12 layers and 12 heads with a hidden dimension of 768. We concatenate the tokenized sequences of the premise sentence and hypothesis sentence with a $[\texttt{SEP}]$ token. 
Because of the size of the rotation matrix, we can't look for distributed representations across all tokens; we look only at the representations of the $[\texttt{CLS}]$ token because the final classification is made from this token's representation in the last layer.

\subsection{High-Level Models} 
We use \methodabbr\ to evaluate whether BERT fine-tuned on MoNLI will represent two boolean intermediate variables. The first is an indicator variable for negation, which is true if and only if negation is present in the premise and hypothesis. The second is a variable that is true if $w_p$ entails $w_h$. This model is perhaps best expressed as a simple program (\Figref{fig:nliprog}). 
Again, we  also consider two alternative high-level models to contextualize our results. One model represents only lexical entailment and not negation. The other represents the identity of the premise word $w_p$.
\input{tables/nli-results.tex}

\subsection{Results} The \intaccmetricabbr\ results achieved by the best alignment for each high-level model can be seen in \Tabref{tab:monliresults}.
There is a perfect alignment between fine-tuned BERT and a symbolic algorithm with variables representing the presence of negation and the lexical entailment relation between $w_p$ and $w_h$. In \Tabref{tab:monliresults}, this is shown by the perfect \intaccmetricabbr\ for layer~9 and intervention size~256, meaning 256 non-standard basis dimensions of the $[\texttt{CLS}]$ token representation in layer~9 of BERT encode the relation between $w_p$ and $w_h$ and 256 other non-standard basis dimensions encode negation. Across all alignments and intervention types, \methodabbr\ learns more faithful alignments (higher \intaccmetricabbr) than a brute-force search through alignments, and no localist alignment comes close to the learned distributed alignments in terms of \intaccmetricabbr. 

However, the distributed representation of the lexical entailment relation between $w_p$ and $w_h$ can be nearly perfectly decomposed into two representations that encode the identity of the word $w_p$ and the identity of the word $w_h$, respectively. This result is shown by the near perfect \intaccmetricabbr\ in the final column of \Tabref{tab:monliresults}. This tells us that what appeared to be a representation of the lexical entailment, was in fact a ``data structure'' of two word identity representations. 

\section{Conclusion}

We introduce \methodname\ (\methodabbr), a method to align interpretable causal variables with distributed neural representations. We learn distributed alignments that are more interpretable than localist alignments and do so with a gradient-descent based search method that improves upon the state-of-the-art brute-force search. In our two experiments, we discovered perfect alignments of distributed neural representations to binary high-level variables encoding simple equality and lexical entailment relations. However, when we investigated the substructure of these representations, we found that the lexical entailment representations could be decomposed into  sub-representations of word identity. This highlights the need to investigate the causal substructure of neural representations. On the other hand, the presence of perfect representations of simple equality relations that cannot be decomposed into representations of the entities in the relations is a foundational result that should inform our understanding of how and when symbolic and connectionist architectures coexist.

\section*{Acknowledgments} 

This research is supported in part by grants from Open Philanthropy, Meta~AI, Amazon, and the Stanford Institute for Human-Centered Artificial Intelligence (HAI).

\bibliography{references}

\newpage
\clearpage

\appendix

\section*{Supplementary Materials}

\section{Experimental Setup Details}\label{app:general-exp-setup}

\subsection{Training Data for \methodname\ (\methodabbr{})}\label{app:training-data-construction}

For each task, we create training datasets for learning the rotation matrix of each high-level model. As defined in \Defref{def:intinv}, each input--output pair for training the rotation matrix consists of a base input that has two pairs of input values. Additionally, we have a set of source inputs mapping to interventions on different intermediate variables, and the corresponding counterfactual outputs (i.e., the updated outputs under interventions). Note that only for cases where there are multiple high-level intermediate variables involved, we sample more than one source input. For such cases, we randomly choose to interchange two variables together from two source inputs or swap a single variable from a single source input.

\paragraph{Hierarchical Equality Experiments} For our high-level models abstracting both equality relations and left equality relation, we sample a set of source inputs and interchange the equality relations of the corresponding shape pairs from the source inputs with the equality relations from the base input. For our high-level model abstracting the identity of the first shape, we sample a source input and interchange the first change from the source input with the base input.

\paragraph{Monotonicity NLI Experiments} For our high-level models abstracting negation or lexical entailment, we sample a set of source inputs and interchange the boolean value for negative or the value for lexical entailment from the source inputs with the base input. For our high-level model abstracting only the identity of replacing lexeme from the \emph{hypothesis} sentence, we sample another \emph{hypothesis} sentence from the one seen in training set and interchange its lexeme with the base input. To avoid cases where entailment labels are invalid (e.g., the entailment relation between ``car'' and ``tree'' is ambiguous), we specifically sample a valid English word that is either a hypernym or a hyponym of the lexeme item in the \emph{premise} sentence, and from a new lexeme pair. Then, we construct a new pair of \emph{premise} and \emph{hypothesis} sentences by sampling a sentence template (i.e., a sentence with replaceable lexeme position such as ``a man is talking to someone in a [$\texttt{lexeme}$]'') from the training dataset and replacing the lexeme items with new ones.
 
\subsection{Reproducibility}\label{app:reproducibility}

\paragraph{Hierarchical Equality Experiment} We randomly generate 1.92M input--output pairs for training the model. We train our model for 10 epochs before reaching 100\% training accuracy for the task. We also evaluate model performance on a hold-out testing set with unseen input-output pairs, and our model achieves 100\% testing accuracy. For each high-level model, we then generate a training dataset for learning the rotation matrix. For each high-level model, we construct 640K such input--output pairs as our training data and 19.2K pairs as our testing data.

For both training phases, we use a batch size of 6.4K with a maximum training epoch of 10. We set the learning rate to 1e$^{-3}$ with an early stop patient step set to 10K. Training with a single NVIDIA 2080 Ti RTX 11GB GPU takes less than ten minutes to converge. All datasets were balanced across the two labels during standard and interchange intervention training objectives. We run each experiment three times with distinct random seeds.

\paragraph{Monotonicity NLI Experiment} We randomly sample 10K examples from the original MoNLI dataset and use it to train our low-level models to solve MoNLI. We finetune our model for 5 epochs before reaching 100\% training accuracy for the task. We also evaluate model performance on a hold-out testing set, and our model achieves 100\% testing accuracy. For training and evaluating the rotation matrix of each high-level model, we create 24K examples as our training dataset for the first high-level model, and 10K for the rest two high-level models. For evaluation, we create 1.92K for the first high-level model, and 1K for the rest two high-level models. 

We finetune our model for 5 epochs with a learning rate of $2e^{-5}$ before reaching 100\% task accuracy with a batch size of 32. For the learning rotation matrix, we use a batch size of 64 with a learning rate of $2e^{-3}$ for a fixed epoch number of 5. Training with a single NVIDIA 2080 Ti RTX 11GB GPU takes less than ten minutes to converge for both training phases. We run each experiment three times with distinct random seeds. 

\subsection{Brute-Force Search Baseline}\label{app:greedy-search}
Without additional training, our brute-force search baseline finds the best \intaccmetricabbr\ by searching over possible alignments $(\Pi, \tau)$ as in Definition~\ref{def:intinvacc}. For simple feed-forward networks, we map a high-level variable to a set of low-level variables within a sliding window with a size equal to the intervention size. We then incrementally search for the sliding window achieving the best \intaccmetricabbr\ score starting from the first index of the intervened representation in the network. For Transformer-based networks, we avoid searching over all possible windows to make computation tractable, by only looking at windows with a starting index from $\{0, 64, 128, 256, 512\}$ of the $[\texttt{CLS}]$ token representation. Instead of targeting a specific set of layers in neural networks, we perform searches over all layers. Note that for the worst-case scenario, the number of hypotheses for the brute-force search approach becomes intractable and can be estimated as $C^{n}_{m}$ where $n$ is the total dimension size of the neural representation, and $m$ is the variable dimension size.

\newpage

\subsection{Localist Alignment Baseline}\label{app:localist}
Without additional training, our localist alignment baseline finds a local optimal localist alignment matrix based on the learned rotation matrix. We pick the rotation matrix with the best \intaccmetricabbr\ result from each category for evaluation. To find a localist alignment matrix, we follow Algorithm~\ref{alg:learning} to get our localist alignment matrix $\mathcal{L}$ from any orthogonal matrix $\mathcal{R}$. We then use $\mathcal{L}$ as our rotation matrix and evaluate \intaccmetricabbr\ following our evaluation paradigm. 

\begin{algorithm}[hp]
   \caption{\textbf{Finding Localist Alignment Matrix}}
\label{alg:learning}
\newcommand{\STATE}{\li }
\begin{codebox}
\Procname{$\proc{FindLocalistAlignment}(\mathcal{R})$}
\STATE \Comment $\mathcal{R}$ is an orthogonal matrix.
\STATE $\mathcal{R}_{\text{a}} = \mathcal{R}.aboslute\_value()$
\STATE $\mathcal{L} = \texttt{torch.zeros\_like}(\mathcal{R})$
\STATE $\mathcal{P} = [ ]$
\STATE \For $i=0; i < \mathcal{R}\texttt{.shape[0]}; i{+}{+}$
\STATE \Do $\mathcal{P} \mathrel{+}=[(\mathcal{R}_{\text{a}}=\texttt{torch.max}(\mathcal{R}_{\text{a}}))\texttt{.nonzero()}]$
\STATE $\mathcal{R}_{\text{a}}[\mathcal{P}[-1]\texttt{.row}, :] = 0.$
\STATE $\mathcal{R}_{\text{a}}[:, \mathcal{P}[-1]\texttt{.col}] = 0.$
\End
\STATE \For $p \in \mathcal{P}$
\STATE \Do $\mathcal{L}[p\texttt{.row}, p.\texttt{col}] = 1.$
\End
\STATE $\mathcal{P} = \mathcal{P} * \texttt{get\_sign}(\mathcal{R})$
\STATE \Return $\mathcal{P}$
\End
\end{codebox}
\end{algorithm}

\newpage

\subsection{Subspace \methodabbr{}}\label{app:subspace-search}

After learning a rotation matrix, we can fix it and learn another rotation matrix on top of it to do subspace high-level variable alignment. For instance, in the case of our MoNLI experiment, we fix the rotation matrix aligning the Lexical Entailment representation and further test whether we can learn another rotation matrix to align word identity. To achieve this, we initialize the first rotation matrix which aligns a larger subspace and freezes its weights along with the rest of the model. Then, we train another rotation matrix by taking the output representations from the first one with the same training objective as the first one as defined in Definition~\ref{def:trainobj}. The training data for the second rotation matrix is not the same as the first one, where we use the training data for the high-level model hypothesized to align with the subspace (e.g., the training data for the identity of first argument for the hierarchical equality task, and the training data for the identity of lexeme for the MoNLI task). Note that for both of our experiments, the subspace dimension is half of its parent subspace for simplicity.

\section{Runtime Comparison: Brute-force Search Baseline vs.~\methodabbr{}}\label{app:runtime-compare}

\Tabref{tab:runtime} shows the runtime comparison between our method and brute-force search under the same settings for each task. Only our approach requires training. We underestimate the runtime for the brute-force search approach by only considering a limited set of possible alignments without exhaustively searching over the entire combination, which leads to intractable computations (See the BFS$_{\text{max}}$ column of \Tabref{tab:runtime}). The runtime of our approach can be further optimized if we deploy early stopping or optimized training data size, and it is invariant with the number of testing hypotheses. 

\input{tables/runtime}

\input{figures/RotationDegree}

\section{Remarks on Learned Rotation Matrix}

\Figref{fig:rotation-degree} shows the rotation in degree(s) of eigenvectors\footnote{The eigenvectors of a rotation matrix are the vectors that remain unchanged after the rotation.} of our learned rotation matrix for each task. We pick the best-performing oracle low-level model for each task for analyses. Our results suggest that learned rotations are not trivial, as the majority of basis vectors are rotated. These results suggest that the representations of high-level variables are highly distributed where direct probes over learned activation may fail to reveal the actual causal role of the representation effectively.

\section{Common Questions}

In this section, we answer common questions that may be raised while reading this report.

\begin{quote}
\emph{Is the learned orthogonal matrix orthonormal?}
\end{quote}

Yes. We use the trainable \texttt{orthogonal} matrix implementation from PyTorch's \texttt{torch.nn.utils.} \texttt{parametrizations}. It guarantees the resulting matrix is orthonormal when the rotation matrix is a full square matrix. Keeping the matrix orthonormal is crucial since it ensures we focus on rotation rather than scaling. Details can be found at \url{https://pytorch.org/docs/stable/generated/torch.nn.utils.parametrizations.orthogonal.html}.

\begin{quote}
\emph{How stable is the optimization process of the orthogonal matrix?}
\end{quote}

We rely on the default initialization of the orthogonal matrix in \texttt{pytorch}. The initialization step is important for finding the local optimal of the rotation matrix. In our experiment, we use random seeds and pick the best results out of our distinct runs to address this issue. However, we may consider different initialization schemes in the future.

\begin{quote}
\emph{Is an orthogonal matrix required to find distributed alignments?}
\end{quote}

In principle, the transformation is not required to be an orthogonal matrix. In fact, an orthogonal matrix assumes a linear transformation before aligning with a high-level variable, which may not be optimal if the aligning variable is represented in a non-linear sub-manifold of the representation space. In such cases, an orthogonal transformation results in imperfect interchange intervention accuracy, and an invertible and differentiable non-linear transformation may be more suitable (e.g., normalizing flow or invertible neural network). In practice, this transformation is computationally difficult to find, and the linear connections within neural networks also make them unlikely to be required to find alignments. We leave these investigations to future works.

\begin{quote}
\emph{What are the prerequisites to deploy this analysis method in practice?}
\end{quote}

We assume a partial or complete causal graph of the data generation process. Specifically, we assume to have interchangeable high-level variables defined for the causal graph. Additionally, we assume we can sample counterfactual data (i.e., base and source inputs where they differ in values of high-level variables) based on the causal graph.

\begin{quote}
\emph{How to interpret the result if the interchange intervention accuracy is not 100\%?}
\end{quote}

When \intaccmetricabbr\ is $\alpha<$100\%, we rely on the graded notion of $\alpha$\textit{-on-average} approximate causal abstraction \cite{geigericard}, which directly coincides with \intaccmetricabbr. More importantly, the relative \intaccmetricabbr\ rankings between the high-level models also show which high-level model is a better approximation of the low-level model.

\begin{quote}
\emph{Does \methodabbr\ scale with large foundation models?}
\end{quote}

Currently, the number of learnable parameters of the rotation matrix groups in polynomial time with the size of hidden representations. For instance, if our intervention site size is 512 in the lower-level model, the number of parameters of the rotation matrix is $512 \times 512$, which is about 0.26M. If we want to rotate concatenated token sequence embeddings of a \texttt{BERT-BASE} model in any layer, the number of parameters of the full rotation matrix is about 15.4B which becomes intractable for standard training infrastructure. To make computation tractable, \methodabbr\ should be further reducible by representing only the aligned subspace, not the full rotation matrix. For instance, to find a 2-dim distributed representation within a 512-dimensional representation space, we approximately only need to learn $512 \times 2$ parameters. In addition, we may use a low-rank approximation of the rotation matrix.

\begin{quote}
\emph{What are some practical usage of \methodabbr{}?}
\end{quote}

Practically, DAS transforms representations into an operatable state where interchange intervention results in interpretable model behaviors. DAS, itself, is a powerful tool for conducting causal abstraction analysis of a neural network.

\section{Task Performance \& Interchange Intervention Accuracy Over Training Epochs}\label{app:results-over-training}

\begin{figure*}[hp]
\centering
\subfigure[$|\mathbf{N}| =16$]{
  \centering
  \includegraphics[width=0.8\linewidth]{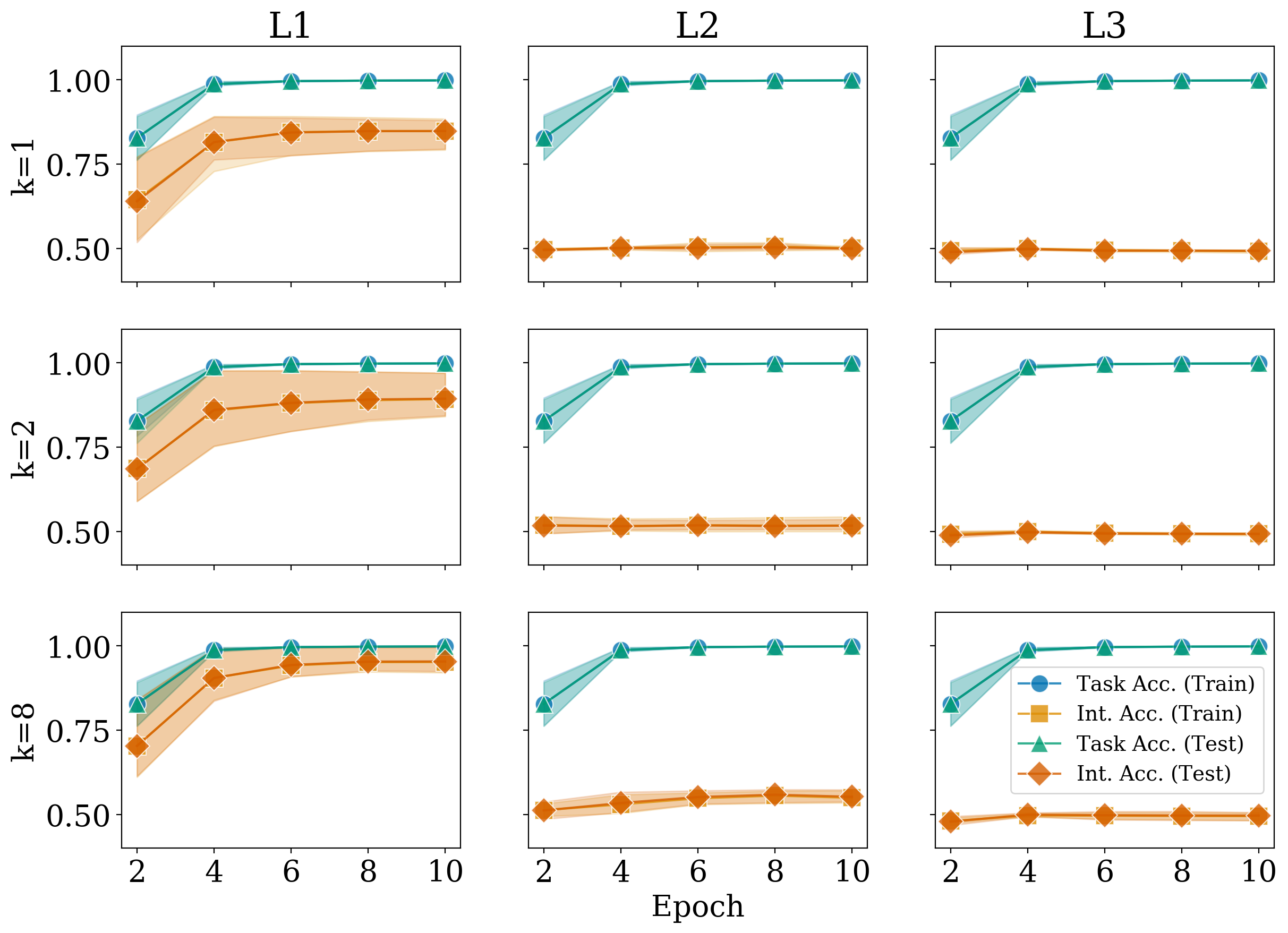}
  \label{fig:sub1}
  }
\subfigure[$|\mathbf{N}| =32$]{
  \centering
  \includegraphics[width=0.8\linewidth]{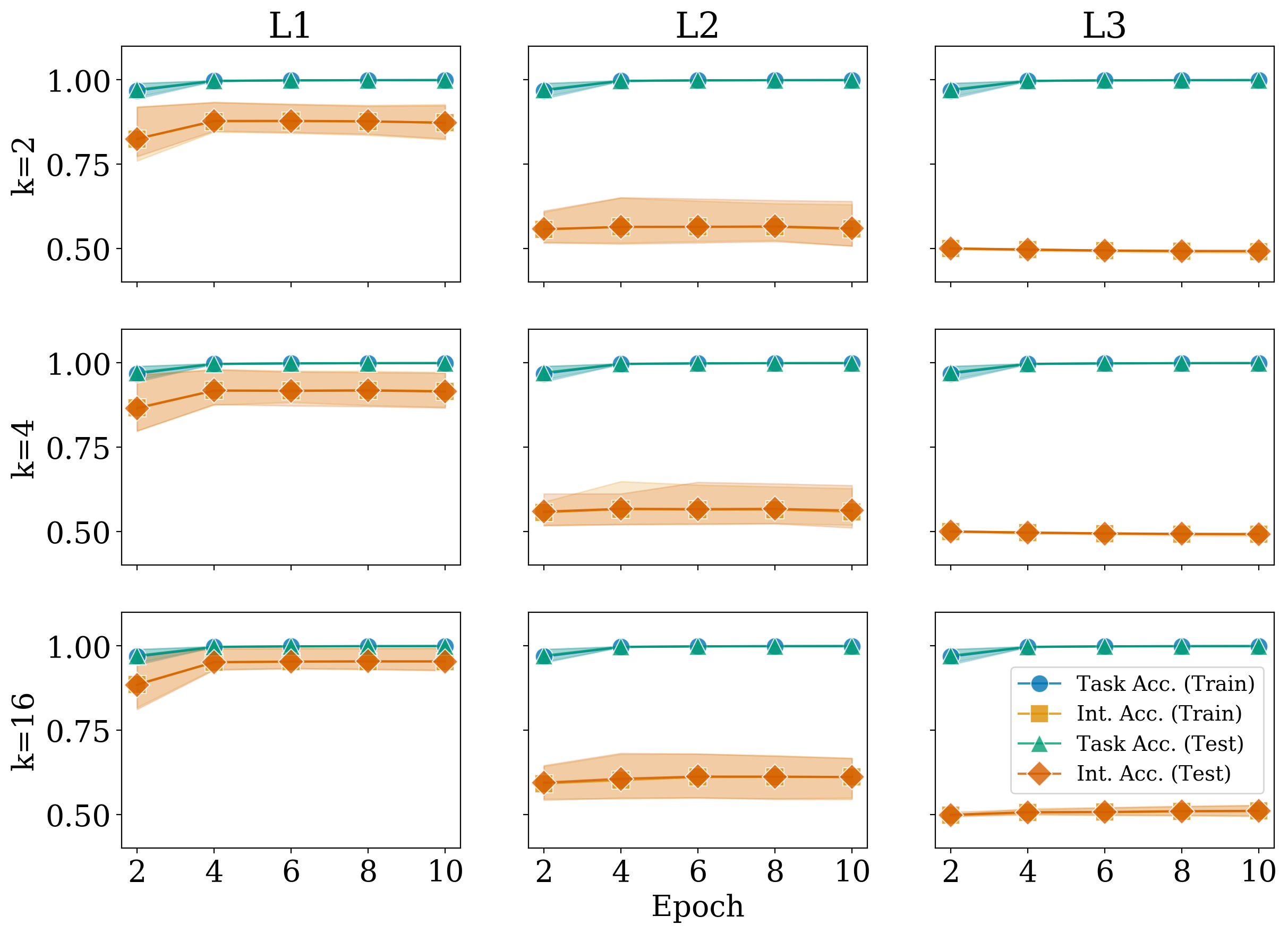}
  \label{fig:sub2}
  }
\caption{Accuracy over training epochs of the high-level model abstracting both equality relations for hierarchical equality experiment.}
\label{fig:equality-1}
\end{figure*}

\begin{figure*}[hp]
\centering
\subfigure[$|\mathbf{N}| =16$]{
  \centering
  \includegraphics[width=0.8\linewidth]{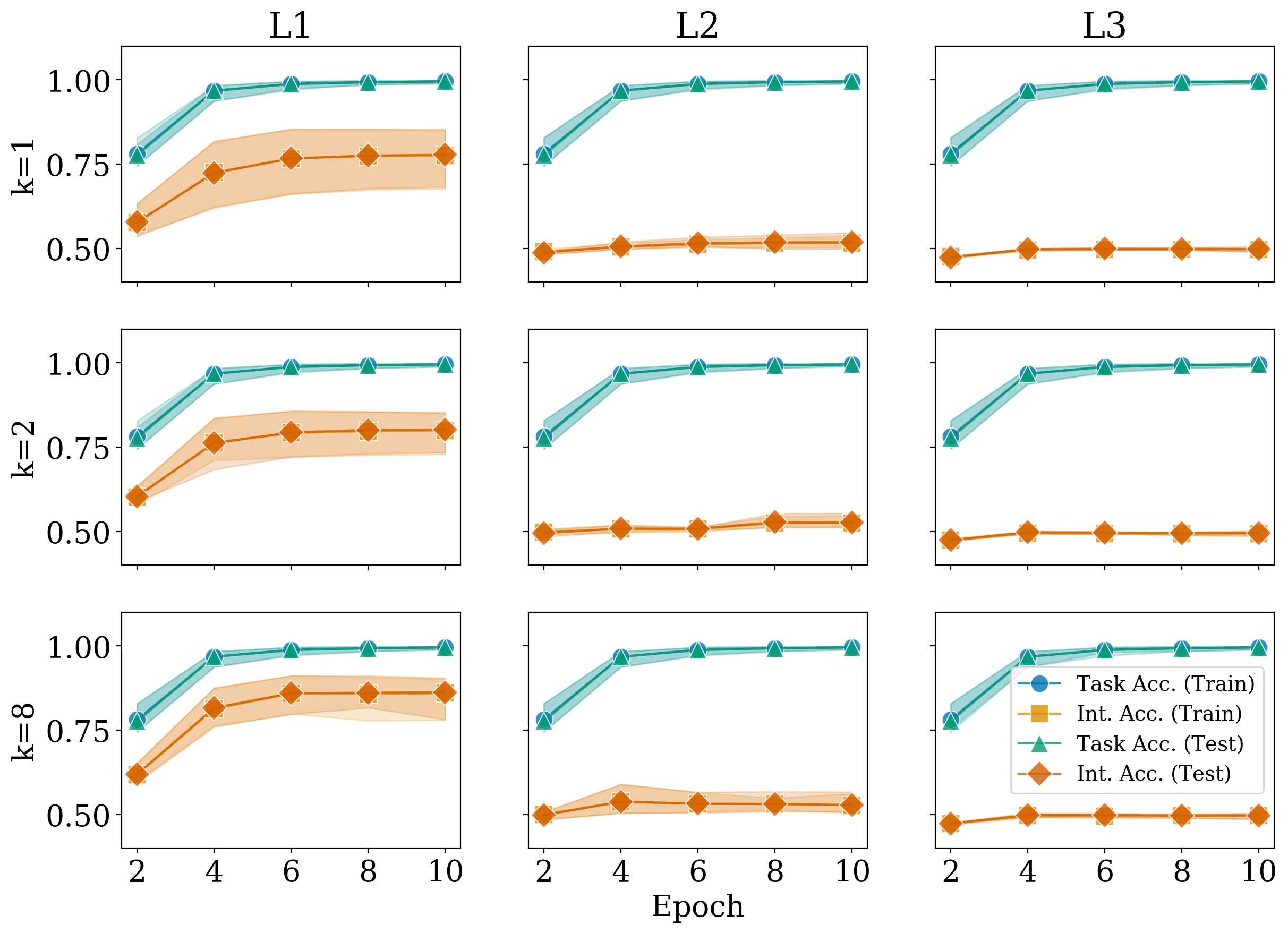}
  \label{fig:sub3}
  }
\subfigure[$|\mathbf{N}| =32$]{
  \centering
  \includegraphics[width=0.8\linewidth]{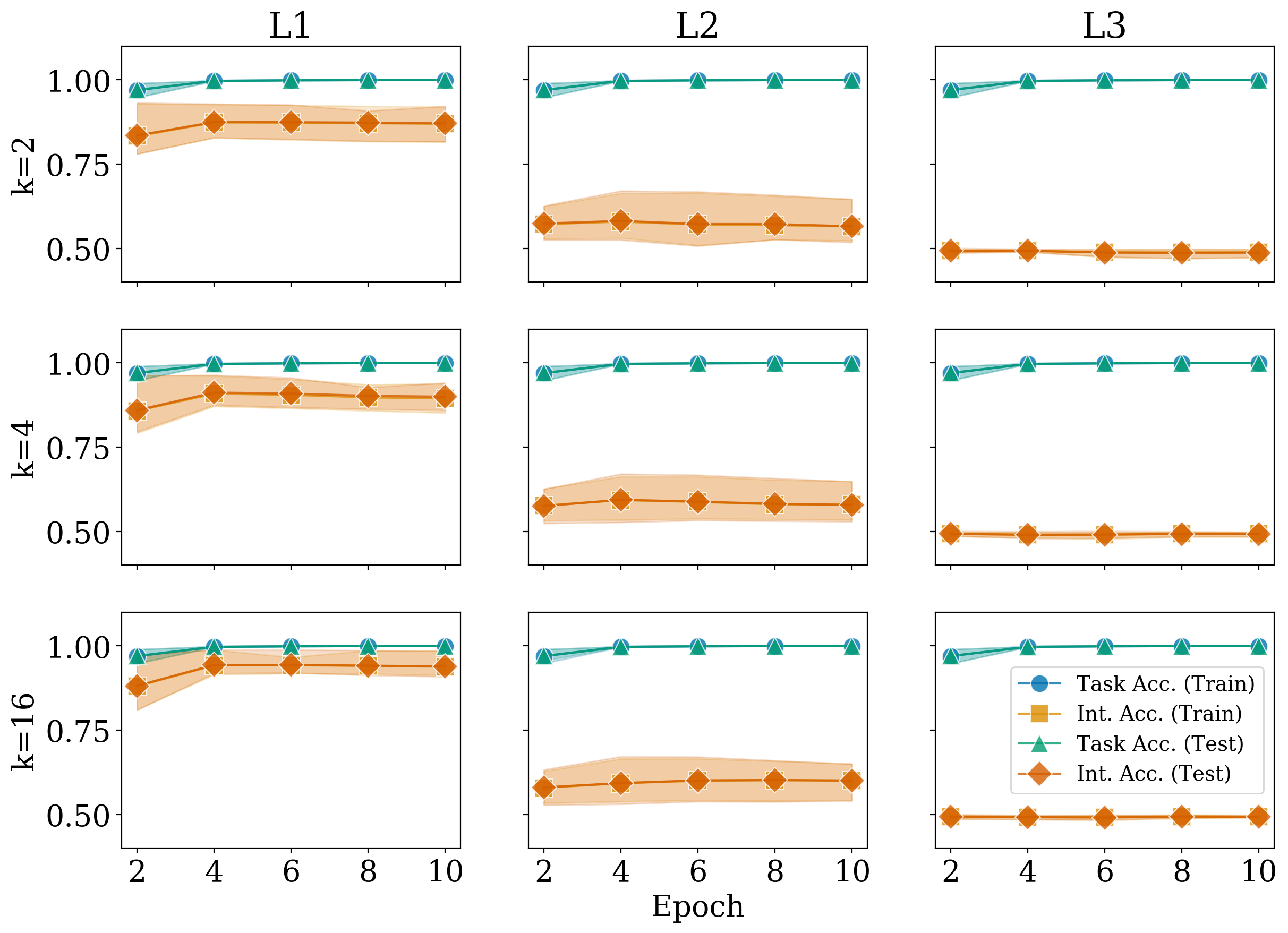}
  \label{fig:sub4}
  }
\caption{Accuracy over training epochs of the high-level model abstracting left equality relation for hierarchical equality experiment.}
\label{fig:equality-2}
\end{figure*}

\begin{figure*}[hp]
\centering
\subfigure[$|\mathbf{N}| =16$]{
  \centering
  \includegraphics[width=0.8\linewidth]{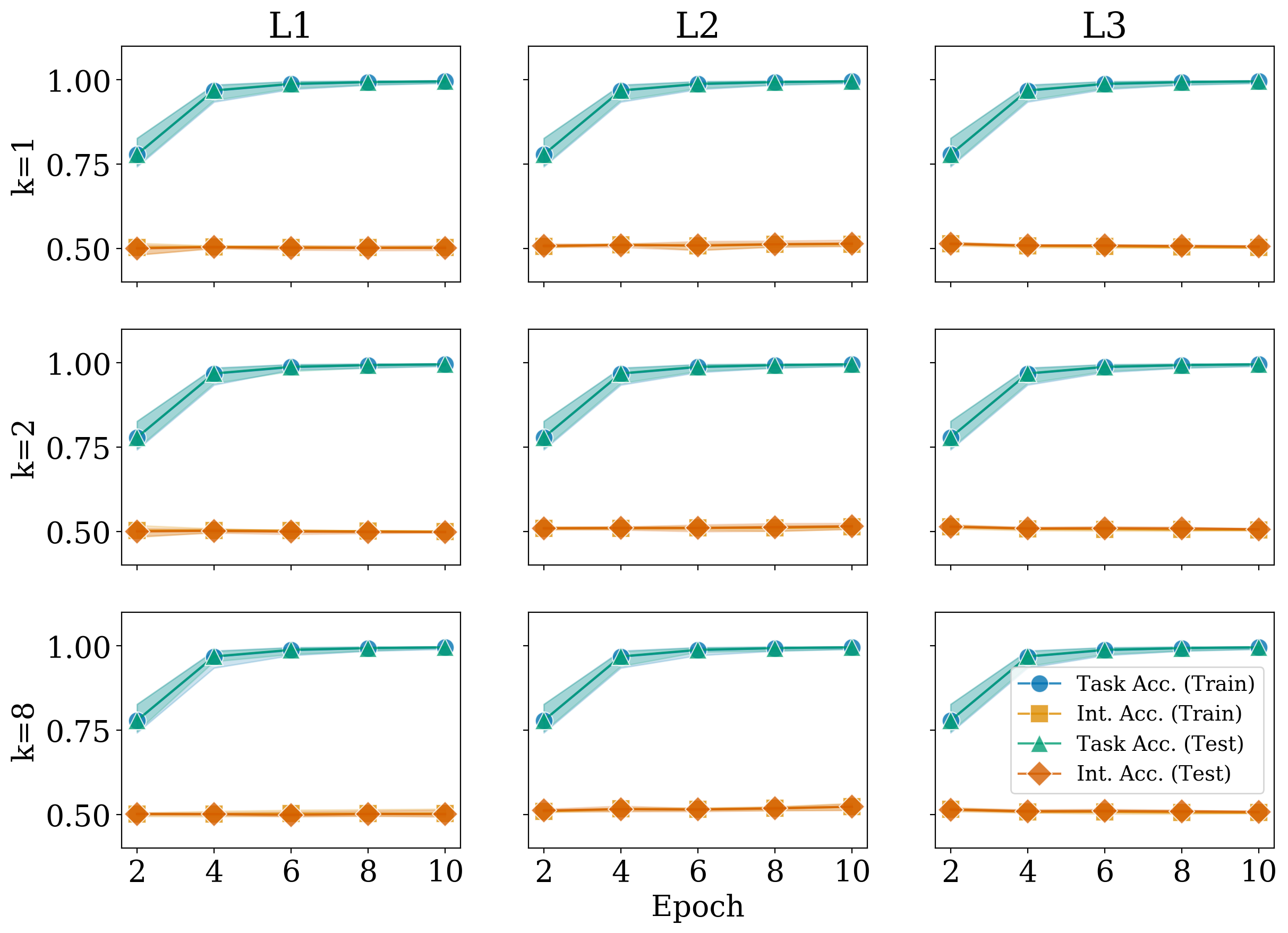}
  \label{fig:sub5}
  }
\subfigure[$|\mathbf{N}| =32$]{
  \centering
  \includegraphics[width=0.8\linewidth]{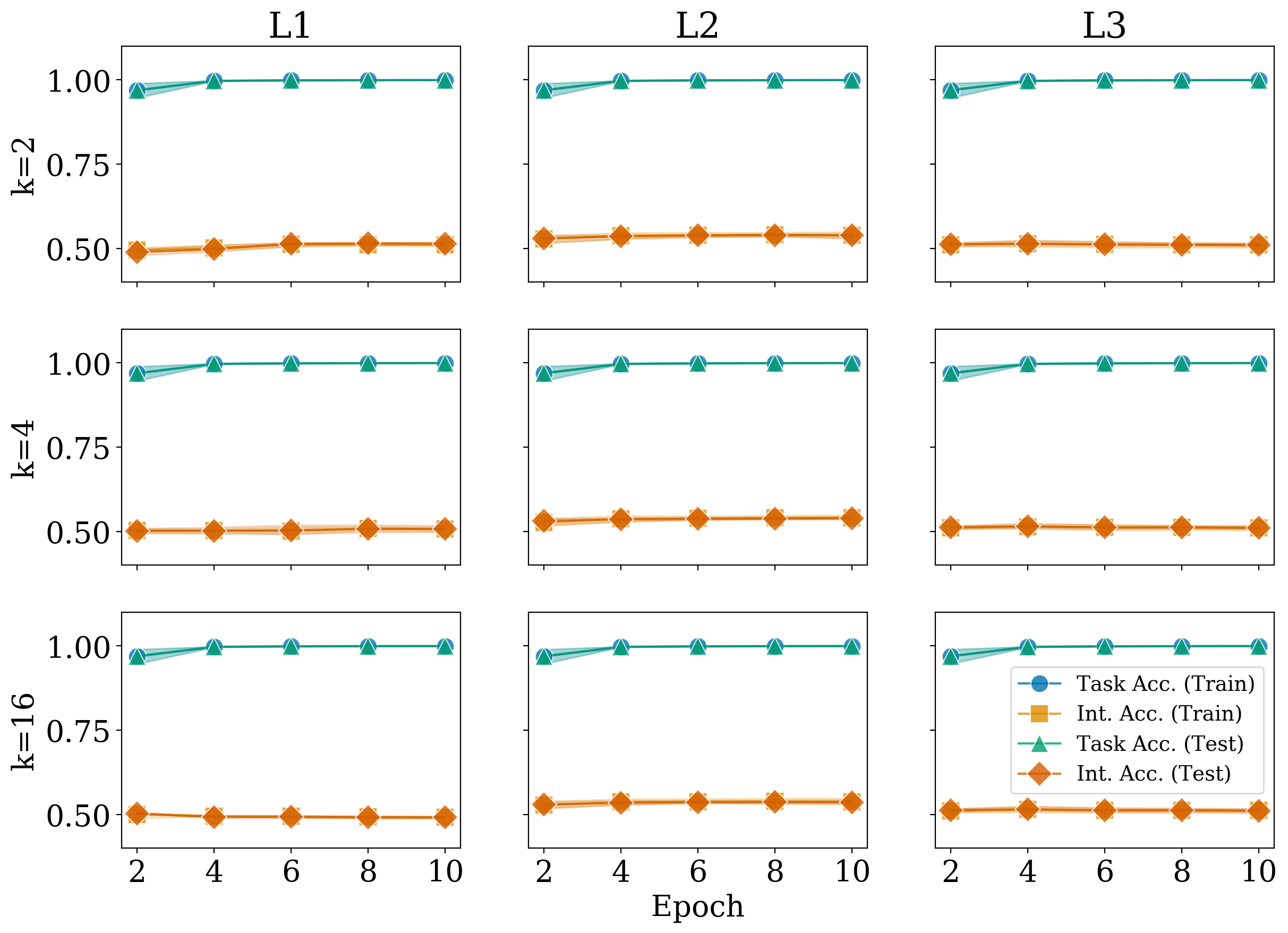}
  \label{fig:sub6}
  }
\caption{Accuracy over training epochs of the high-level model abstracting identity of first argument for hierarchical equality experiment.}
\label{fig:equality-3}
\end{figure*}

\begin{figure*}[hp]
    \centering
    \includegraphics[width=0.5\textwidth]{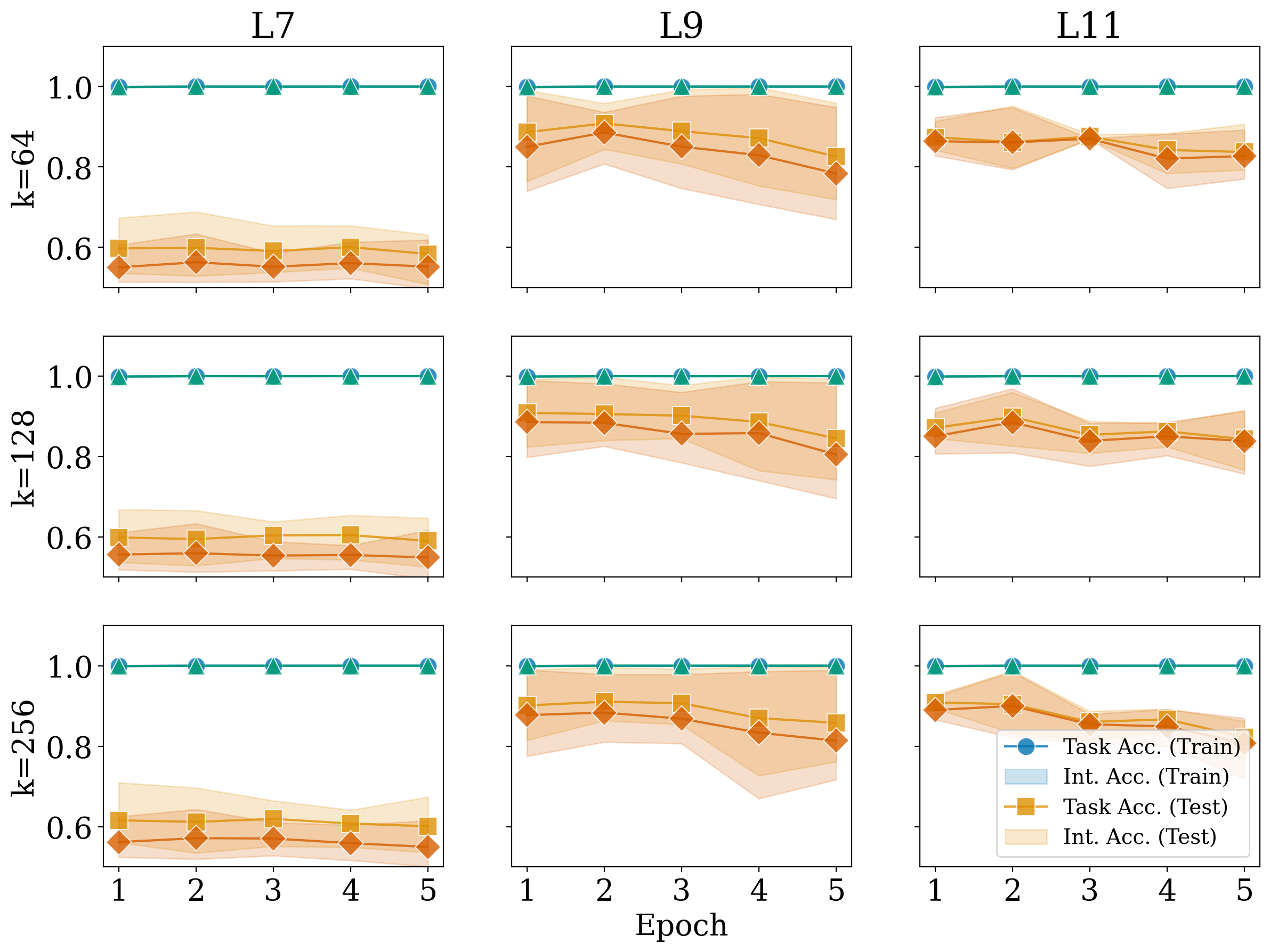}
    \caption{Accuracy over training epochs of the high-level model abstracting both negative and lexical entailment with $|\mathbf{N}| =768$ for monotonicity NLI experiment.}
    \label{fig:monli-1}
\end{figure*}

\begin{figure*}[hp]
    \centering
    \includegraphics[width=0.5\textwidth]{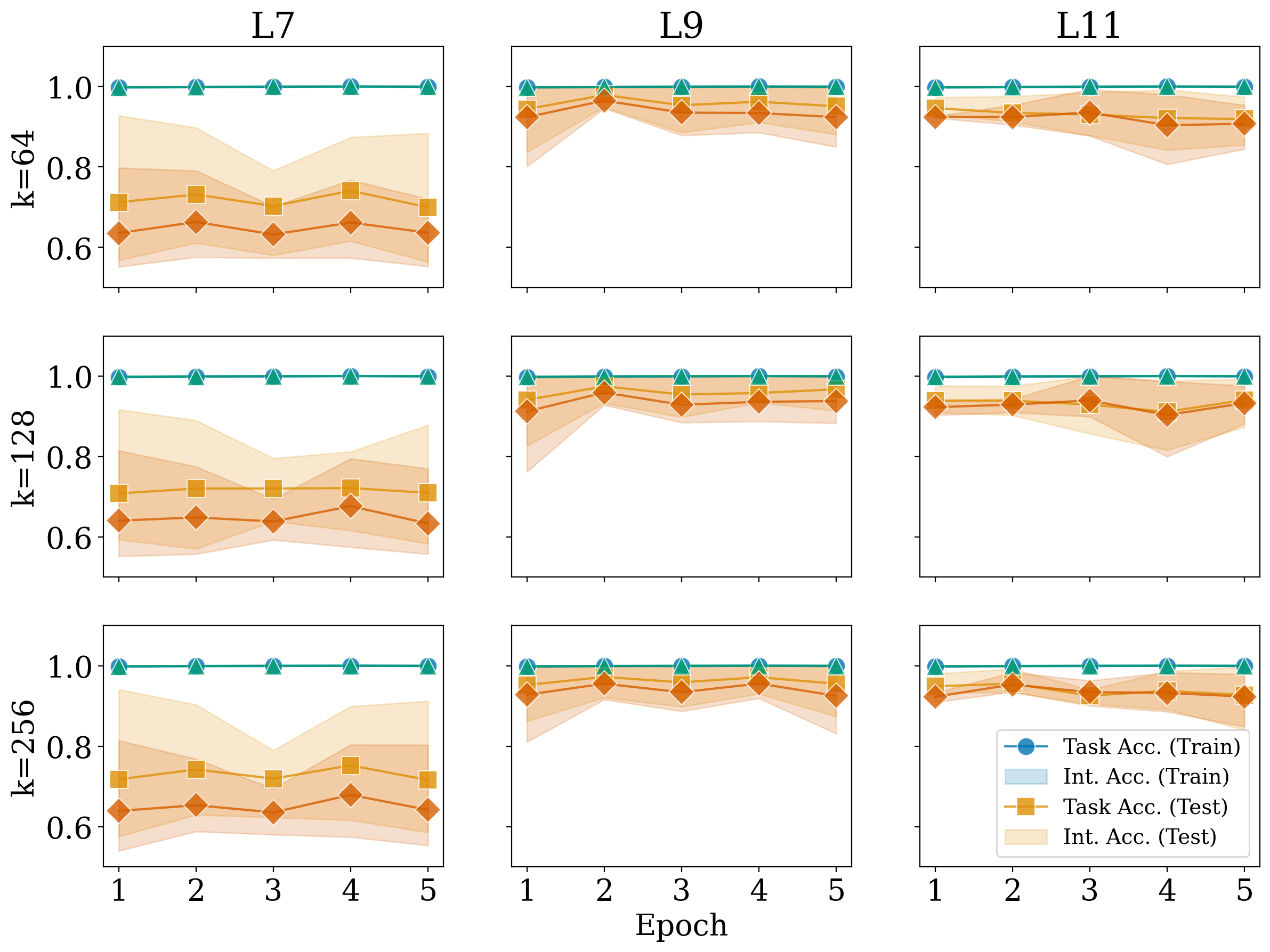}
    \caption{Accuracy over training epochs of the high-level model abstracting lexical entailment with $|\mathbf{N}| =768$ for monotonicity NLI experiment.}
    \label{fig:monli-2}
\end{figure*}

\begin{figure*}[hp]
    \centering
    \includegraphics[width=0.5\textwidth]{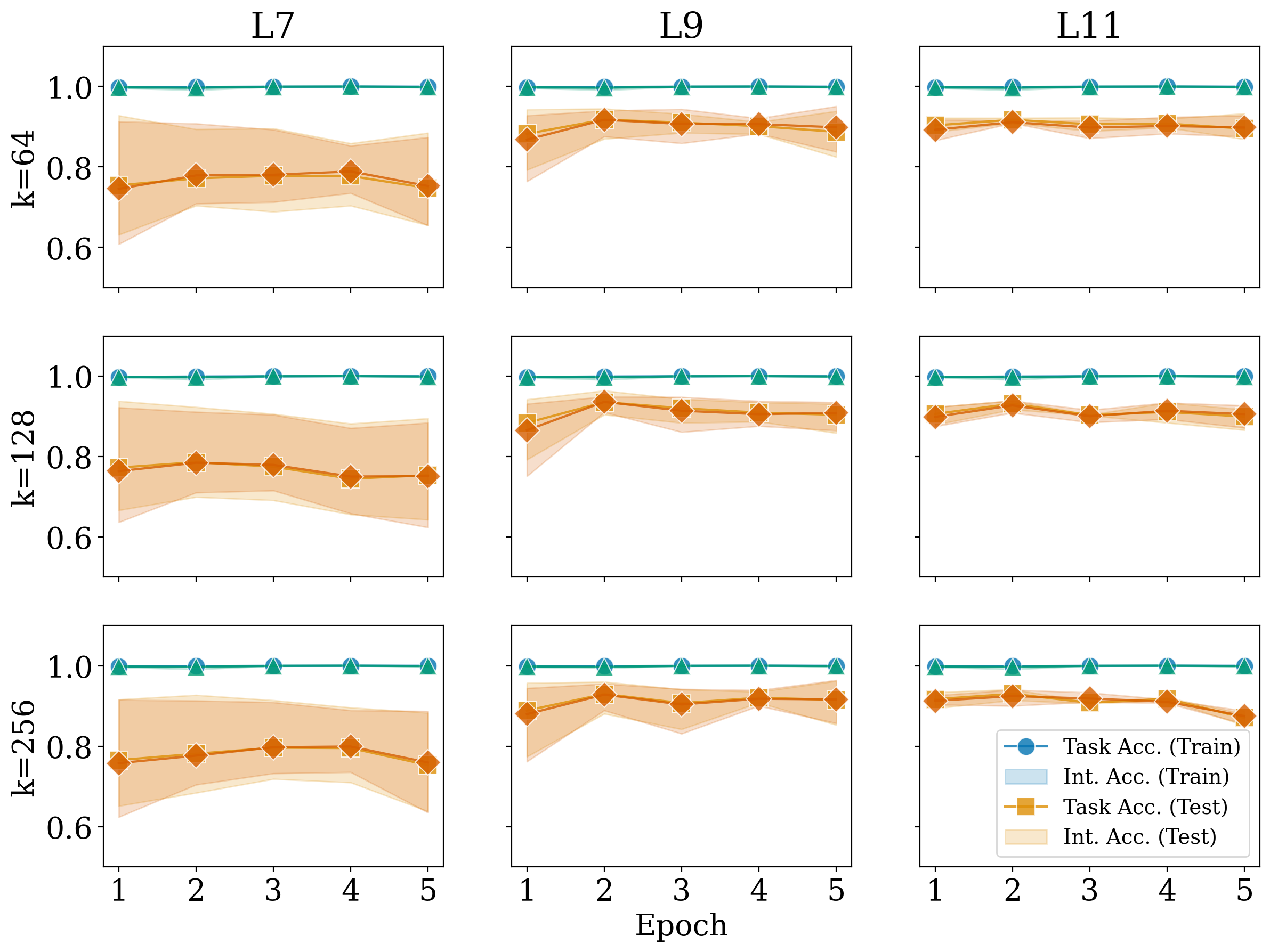}
    \caption{Accuracy over training epochs of the high-level model abstracting the identity of lexeme with $|\mathbf{N}| =768$ for monotonicity NLI experiment.}
    \label{fig:monli-3}
\end{figure*}

We additionally measure task performance (Task Acc.)\ as well \intaccmetricabbr\ (Int.~Acc.)\ of our alignments over training epochs for both seen training examples as well as unseen testing examples. Our results are shown from \Figref{fig:equality-1} to \Figref{fig:monli-3}.

\end{document}

%% file: methods-alt.tex
\tikzset{
  state/.style={
    fill=gray!80,
    draw,
    font=\scriptsize,
    shape=rectangle,
    minimum width=1cm,
    minimum height=0.5cm
  },
  rep/.style={
    fill=blue!20,
    draw,
    font=\scriptsize,
    shape=rectangle,
    minimum width=2cm,
    minimum height=0.5cm
  },
  align/.style={
    dashed,
    ->
  },
  intervention/.style={
    color=orange,
    thick,
    ->
  }
}

\newcommand{\sources}{ \{\mathbf{s}_j\}_1^k}
\newcommand{\sites}{ \{\mathbf{X}_j\}_1^k}
\newcommand{\rotdeg}{20^{\circ}}
\newcommand{\distint}[5]{\mathbf{#1} \overset{#4}{\leftarrow} \mathbf{#5}}
\newcommand{\rmat}{\mathbf{R}}

\newcommand{\booleanNetworkbNoInput}[4]{
  \node(x01) at (#1, #2){};
  \node[rep,fill=orange!20, above=0.2cm of x01](h01){$H_{1} = 0.6$};
  \node[rep,fill=orange!20,  right=0.2cm of h01](h11){$H_{2} = 1.28$};
  \node[rep,fill=orange!20, above=0.2cm of h01, xshift=1.1cm](y1){$O = 0.08$};
  \node[above=0.2cm of y1](lol){$\True$};
  \path(h01.north) edge[] (y1.south);
  \path(h11.north) edge[] (y1.south);
}

\newcommand{\booleanNetworkccNoOutput}[4]{
  \node[rep, fill=yellow!20](x0#2) at (#1, 0){$0$};
  \node[rep, right=0.2cm of x0#2, fill=yellow!20](x1#2){$1$};
  \node[rep, above=0.2cm of x0#2, fill=yellow!20](h0#2){$H_{1} = -0.34$};
  \node[rep, right=0.2cm of h0#2, fill=yellow!20](h1#2){$H_{2} = 0.94$};
  \path(x0#2.north) edge[] (h0#2.south);
  \path(x0#2.north) edge[] (h1#2.south);
  \path(x1#2.north) edge[] (h0#2.south);
  \path(x1#2.north) edge[] (h1#2.south);
}

\newcommand{\booleanModelb}[4]{
  \node[state](x0#2) at (#1, 0){$\False$};
  \node[state, right=0.2cm of x0#2](x1#2){$\True$};
  \node[state,fill=orange!20, above=0.2cm of x0#2](h0#2){$V_{1} = \True$};
  \node[state, right=0.2cm of h0#2](h1#2){$V_{2} = \True$};
  \node[state,fill=orange!20, above=0.2cm of h0#2, xshift=0.6cm](y#2){$V_{3} = \True$};
  \path(x0#2.north) edge[] (h0#2.south);
  \path(x1#2.north) edge[] (h1#2.south);
  \path(h0#2.north) edge[] (y#2.south);
  \path(h1#2.north) edge[] (y#2.south);
}

\newcommand{\booleanModelc}[4]{
  \node[state](x0#2) at (#1, 0){$\True$};
  \node[state, right=0.2cm of x0#2](x1#2){$\True$};
  \node[state, above=0.2cm of x0#2](h0#2){$V_{1} = \True$};
  \node[state, right=0.2cm of h0#2](h1#2){$V_{2} = \True$};
  \node[state, above=0.2cm of h0#2, xshift=0.6cm](y#2){$V_{3} = \True$};
  \path(x0#2.north) edge[] (h0#2.south);
  \path(x1#2.north) edge[] (h1#2.south);
  \path(h0#2.north) edge[] (y#2.south);
  \path(h1#2.north) edge[] (y#2.south);
}

\newcommand{\booleanModelO}[4]{
  \node[state](x0#2) at (#1, 0){$P$};
  \node[state, right=0.2cm of x0#2](x1#2){$Q$};
  \node[state, above=0.2cm of x0#2](h0#2){$V_{1} = p$};
  \node[state, right=0.2cm of h0#2](h1#2){$V_{2} = q$};
  \node[state, above=0.2cm of h0#2, xshift=0.6cm](y#2){$V_{3} = v_{1} \land v_{2}$};
  \path(x0#2.north) edge[] (h0#2.south);
  \path(x1#2.north) edge[] (h1#2.south);
  \path(h0#2.north) edge[] (y#2.south);
  \path(h1#2.north) edge[] (y#2.south);
}

\newcommand{\booleanNetworkb}[4]{
  \node[rep](x0#2) at (#1, 0){$0$};
  \node[rep, right=0.2cm of x0#2](x1#2){$1$};
  \node[rep,fill=orange!20, above=0.2cm of x0#2](h0#2){$H_{1} = 0.6$};
  \node[rep, right=0.2cm of h0#2](h1#2){$H_{2} = 0.94$};
  \node[rep,fill=orange!20, above=0.2cm of h0#2, xshift=1.1cm](y#2){$O = -0.26$};
  \path(x0#2.north) edge[] (h0#2.south);
  \path(x0#2.north) edge[] (h1#2.south);
  \path(x1#2.north) edge[] (h0#2.south);
  \path(x1#2.north) edge[] (h1#2.south);
  \path(h0#2.north) edge[] (y#2.south);
  \path(h1#2.north) edge[] (y#2.south);
    \node[above=0.2cm of y#2](lol){$\False$};
}

\newcommand{\booleanNetworkc}[4]{
  \node[rep](x0#2) at (#1, 0){$1$};
  \node[rep, right=0.2cm of x0#2](x1#2){$1$};
  \node[rep, above=0.2cm of x0#2](h0#2){$H_{1} = 0.6$};
  \node[rep, right=0.2cm of h0#2](h1#2){$H_{2} = 1.28$};
  \node[rep, above=0.2cm of h0#2, xshift=1.1cm](y#2){$O = 0.08$};
  \path(x0#2.north) edge[] (h0#2.south);
  \path(x0#2.north) edge[] (h1#2.south);
  \path(x1#2.north) edge[] (h0#2.south);
  \path(x1#2.north) edge[] (h1#2.south);
  \path(h0#2.north) edge[] (y#2.south);
  \path(h1#2.north) edge[] (y#2.south);
}

\newcommand{\booleanNetworkO}[4]{
  \node[rep](x0#2) at (#1, 0){$X_1$};
  \node[rep, right=0.2cm of x0#2](x1#2){$X_2$};
  \node[rep, above=0.2cm of x0#2](h0#2){$H_{1} = [x_1; x_2]W_{1}$};
  \node[rep, right=0.2cm of h0#2](h1#2){$H_{2} = [x_1; x_2]W_{2}$};
  \node[rep, above=0.2cm of h0#2, xshift=1.1cm](y#2){$O = [h_{1}; h_{2}]\mathbf{w} + b$};
  \path(x0#2.north) edge[] (h0#2.south);
  \path(x0#2.north) edge[] (h1#2.south);
  \path(x1#2.north) edge[] (h0#2.south);
  \path(x1#2.north) edge[] (h1#2.south);
  \path(h0#2.north) edge[] (y#2.south);
  \path(h1#2.north) edge[] (y#2.south);
}

\newcommand{\True}{\textsc{t}}
\newcommand{\False}{\textsc{f}}
\newcommand{\conjMod}{\mathcal{B}}
\newcommand{\conjNetwork}{\mathcal{N}}

We focus on acyclic causal models \citep{pearl,Spirtes} and seek to provide an intuitive overview of our method. An \textbf{acyclic causal model} consists of input, intermediate, and output \textbf{variables}, where each variable has an associated set of \textbf{values} it can take on and a \textbf{causal mechanism} that determine the value of the variable based on the value of its causal parents. For a simple running example, we modify the boolean conjunction models of \cite{geiger-etal-2021-iit} to reveal key properties of \methodabbr. A causal model $\conjMod$ for this problem can be defined as below, where the inputs and outputs are booleans $\True$ and $\False$. Alongside $\conjMod$, we also define a causal model $\conjNetwork$ of a linear feed-forward neural network that solves the task. Here we show $\conjMod$, $\conjNetwork$, and the parameters of $\conjNetwork$:

{\centering
\resizebox{0.4\textwidth}{!}{
  \begin{tikzpicture}
    \booleanModelO{0}{0}{p}{q}
    \booleanNetworkO{4}{1}{x_{1}}{x_{2}}

    \path(x00.south) edge[align, bend right=12pt,->] (x01.south);
    \path(x10.south) edge[align, bend right=12pt,->] (x11.south);
    \path(h00.north) edge[align, bend left=12pt,->] (h01.north);
    \path(h10.north) edge[align, bend left=12pt,->] (h11.north);
    \path(y0.north) edge[align, bend left=12pt,->] (y1.north);
  \end{tikzpicture}
}
\hfill
\begin{tabular}{cc}
 $ W_{1} = \left[
    \begin{array}{rr}
      \cos(\rotdeg) & -\sin(\rotdeg)
    \end{array}
\right]$ &$\mathbf{w} = \left[\begin{array}{ll} 1 & 1 \end{array}\right]$\\
 $W_{2} = \left[
    \begin{array}{rr}    
      \sin (\rotdeg) & \phantom{-}\cos(\rotdeg)
    \end{array}
  \right]$ &$b = -1.8$\\
  \\
  \\
\end{tabular}
\vspace{-1em}
}

\noindent The model $\mathcal{N}$ predicts $\True$ if $O>0$ and $\False$ otherwise. This network solves the boolean conjunction problem perfectly in that all pairs of input boolean values are mapped to the intended output. 

An input $\mathbf{x}$ of a model $\mathcal{M}$ determines a unique total setting $\mathcal{M}(\mathbf{x})$ of all the variables in the model. The inputs are fixed to be $\mathbf{x}$ and the causal mechanisms of the model determine the values of the remaining variables. We denote the values that $\mathcal{M}(\mathbf{x})$ assigns to the variable or variables $\mathbf{Z}$ as $\textsc{GetValues}_{\mathbf{Z}}(\mathcal{M}(\mathbf{x}))$. For example, $\textsc{GetValues}_{V_{3}}(\conjMod([\True, \False])) = \False$.

\subsection{Interventions}

Interventions are a fundamental building block of causal models, and of causal abstraction analysis in particular. An intervention $\mathbf{I} \leftarrow \mathbf{i}$ is a setting $\mathbf{i}$ of variables $\mathbf{I}$. Together, an intervention and an input setting $\mathbf{x}$ of a model $\mathcal{M}$ determine a unique total setting that we denote as $\mathcal{M}_{\mathbf{I} \leftarrow \mathbf{i}}(\mathbf{x})$.  The inputs are fixed to be~$\mathbf{x}$, and the causal mechanisms of the model determine the values of the non-intervened variables, with the intervened variables $\mathbf{I}$ being fixed to $\mathbf{i}$.

We can define interventions on both our causal model $\conjMod$ and our neural model $\conjNetwork$. For example, $\conjMod_{V_{1} \leftarrow \True}([\False, \True])$ is our boolean model when it processes input $[\False, \True]$ but with variable $V_{1}$ set to $\True$. This has the effect of changing the output value to $\True$. Similarly, whereas $\conjNetwork([0, 1])$ leads to an intermediate values $h_{1} = -0.34$ and $h_2=0.94$ and output value $-1.2$, if we compute $\conjNetwork_{h_{1} \leftarrow 1.34}([0, 1])$, then the output value is $0.48$. This has the effect of changing the predicted value to $\True$, because $0.48 > 0$.

\subsection{Alignment}

In causal abstraction analysis, we ask whether a specific low-level model like $\conjNetwork$ implements a high-level algorithm like $\conjMod$. This is always relative to a specific \textit{alignment} of variables between the two models. An alignment $\Pi = (\{\Pi_X\}_{X}, \{\tau_X\}_{X})$ assigns to each high-level variable $X$ a set of low-level variables $\Pi_{X}$ and a function $\tau_X$ that maps from values of the low-level variables in $\Pi_X$ to values of the aligned high-level variable $X$. One possible alignment between $\conjMod$ and $\conjNetwork$ is shown in the diagram above: $\Pi$ is depicted by the dashed lines connecting $\mathcal{B}$ and $\mathcal{N}$.

We immediately know what the functions for high-level input and output variables are. For the inputs, $\True$ is encoded as $1$ and $\False$ is encoded as $0$, meaning $\tau_{P}(1)= \tau_{Q}(1) = \True$ and $\tau_{P}(0) =\tau_{Q}(0)  = \False$. For the output, the network only predicts $\True$ if $y>0$, meaning $\tau_{V_{3}}(x) = \True$ if $x > 0$, else $\False$. This is simply a consequence of how a neural network is used and trained. The functions for high-level intermediate variables $\tau_{V_{1}}(x)$ and $\tau_{V_{2}}(x)$ must be discovered and verified experimentally.

\subsection{Constructive Causal Abstraction}

Relative to an alignment like this, we can define abstraction:
\begin{definition}{\textnormal{(Constructive Causal Abstraction)}}\label{def:constructive-abstraction}
A high-level causal model $\mathcal{H}$ is a constructive abstraction of a low-level causal model $\mathcal{L}$ under alignment $\Pi$ exactly when the following holds for every low-level input setting $\mathbf{x}$ and low-level intervention $\mathbf{I} \leftarrow \mathbf{i}$:
\[
  \tau(\mathcal{L}_{\mathbf{I} \leftarrow \mathbf{i}}(\mathbf{x})\big )
  =
  \mathcal{H}_{\tau(\mathbf{I} \leftarrow \mathbf{i})}(\tau(\mathbf{x}))
\]
\end{definition}
$\mathcal{H}$ being a causal abstraction of $\mathcal{L}$ under $\Pi$ guarantees that the causal mechanism for each high-level variable $X$ is a faithful rendering of the causal mechanisms for the low-level variables in $\Pi_X$.

\newcommand{\defequal}{\overset{\textit{def}}{=}}

To assess the degree to which a high-level model is a constructive causal abstraction of a low-level model, we perform interchange interventions:
\begin{definition}{\textnormal{(Interchange Interventions)}}\label{def:intinv}
  Given source input settings $\sources$, and non-overlapping sets of intermediate variables $\sites$ for model $\mathcal{M}$, define the \textit{interchange intervention} as the model 
  \[
\intinv(\mathcal{M}, \sources, \sites) =  \mathcal{M}_{\bigwedge^{k}_{j=1} \langle \mathbf{X}_j \leftarrow \mathsf{GetVals}_{\mathbf{X}_j}(\mathcal{M}(s_j))\rangle 
      }
    \]
    where $\bigwedge^{k}_{j=1} \langle \cdot \rangle$ concatenates a set of interventions. %
\end{definition}
A \textit{base} input setting can be fed into the resulting model to compute the counterfactual output value. Consider the following interchange intervention:
\[
\intinv(\conjMod, \{[\True,\True]\}, \{\{V_1\}\}) = 
\conjMod_{\{V_{1}\}\leftarrow\mathsf{GetVals}_{\{V_1\}}(\conjMod({[\True,\True]}))}
\]
We process a base input and a source input, and then we intervene on a target variable, replacing it with the value obtained by processing the source.
Our causal model is fully known, and so we know ahead of time that this interchange intervention yields~$\True$. For our neural network, the corresponding behavior is not known ahead of time. The interchange intervention corresponding to the above (according to the alignment we are exploring) is as follows
\[
\intinv(\conjNetwork, \{[1,1]\}, \{\{H_1\}\}) =
\conjNetwork{\{V_{1}\}\leftarrow\mathsf{GetVals}_{\{H_1\}}(\conjNetwork({[1,1]}))}
\]
And, indeed, the counterfactual behavior of the model and the network $\conjNetwork$ are unequal:
\begin{center}
\resizebox{0.45\textwidth}{!}{
  \begin{tikzpicture}
    \booleanModelb{0}{0}{\True}{\False}
    \booleanModelc{4}{1}{\False}{\True}
    \path(h01.north) edge[intervention, bend right=20pt] (h00.north);
  \end{tikzpicture}
  }
  \;\;\;\;
\resizebox{0.45\textwidth}{!}{
  \begin{tikzpicture}
    \booleanNetworkb{0}{0}{1}{0}
    \booleanNetworkc{4.5}{1}{0}{1}
    \path(h01.north) edge[intervention, bend right=15pt] (h00.north);
  \end{tikzpicture}
  }
\end{center}

\noindent Under the given alignment, the interchange interventions at the low and high level have different effects. Thus, we have a counterexample to constructive abstraction as given in \defref{def:constructive-abstraction}. Although $\conjNetwork$ has perfect behavioral accuracy, its accuracy under the counterfactuals created by our interventions is not perfect, and thus $\conjMod$ is not a constructive abstraction of $\conjNetwork$ under this alignment.

\subsection{Distributed Interventions}

The above conclusion is based on the kind of localist causal abstraction explored in the literature to date. As  noted in \secref{sec:intro}, there are two risks associated with this conclusion: (1) we may have chosen a suboptimal alignment, and (2) we may be wrong to assume that the relevant structure will be encoded in the standard basis we have implicitly assumed throughout.

If we simply rotate the representation $[H_{1}, H_{2}]$ by $-\rotdeg$ to get a new representation $[Y_1, Y_2]$, then the resulting network has perfect behavioral and counterfactual accuracy when we align $V_1$ and $V_2$ with $Y_1$ and $Y_2$. What this reveals is that there is an alignment, but not in the basis we chose. Since the choice of basis was arbitrary, our negative conclusion about the causal abstraction relation was spurious.

This rotation localizes the information about the first and second argument into separate dimensions. To understand this, observe that the weight matrix of the linear network rotates a two dimensional vector by $20^{\circ}$ and the rotation matrix rotates the representation by $340^{\circ}$. The two matrices are inverses. Because this network is linear, there is no activation function and so rotating the hidden representation ``undoes'' the transformation of the input by the weight matrix. Under this non-standard basis, the first hidden dimension is equal to the first input argument and the second hidden dimension is equal to the second input argument.

This reveals an essential aspect of distributed neural representations: there is a many-to-many mapping between neurons and concepts, and thus multiple high-level causal variables might be encoded in structures from overlapping groups of neurons \citep{PDP1,PDP2}. In particular, \citet{Smolensky1986} proposes that viewing a neural representation under a basis that is not aligned with individual neurons can reveal the interpretable distributed structure of the neural representations. 

\input{figures/EqualityDistInterchange}
To make good on this intuition we define a distributed intervention, which first transforms a set of variables to a vector space, then does interchange on orthogonal sub-spaces, before transforming back to the original representation space.
\begin{definition}{\textnormal{Distributed Interchange Interventions}}
We begin with a causal model $\mathcal{M}$ with input variables $\mathbf{S}$ and \textit{source} input settings $\{\mathbf{s}_j\}_{j=1}^k$.
Let $\mathbf{N}$ be a subset of variables in $\mathcal{M}$, the \emph{target variables}. 
Let $\mathbf{Y}$ be a vector space with subspaces $\{\mathbf{Y}_j\}_{0}^{k}$ that form an orthogonal decomposition, i.e., $\mathbf{Y}=\bigoplus_{j=0}^{k} \mathbf{Y}_j$.
Let $\mathbf{R}$ be an invertible function $\mathbf{R}: \mathbf{N} \to \mathbf{Y}$. 
Write $\mathsf{Proj}_{\mathbf{Y}_j}$ for the orthogonal projection operator of a vector in $\mathbf{Y}$ onto subspace $\mathbf{Y}_j$.\footnote{Thus, $\mathsf{Proj}$ generalizes $\mathsf{GetVals}$ to arbitrary vector spaces.}
A \textbf{distributed interchange intervention} yields a new model 
$\dintinv(\mathcal{M}, \rmat, \sources, \{\mathbf{Y}_j\}_{0}^{k})$
which is identical to $\mathcal{M}$ except that the mechanisms $F_{\mathbf{N}}$ (which yield values of $\mathbf{N}$ from a total setting) are replaced by:
\[
F^{*}_{\mathbf{N}}(\mathbf{v}) = \mathbf{R}^{-1}\bigg(
\mathsf{Proj}_{\mathbf{Y}_0}\Big(\mathbf{R}\big(F_\mathbf{N}(\mathbf{v})\big)\Big)\\
+\sum_{j=1}^{k} \mathsf{Proj}_{\mathbf{Y}_j}\Big(\mathbf{R}\big(F_\mathbf{N}(\mathcal{M}(\mathbf{s_j}))\big)\Big)\bigg).
\]
\end{definition}
Notice that in this definition the base setting is partially preserved through the intervention (in subspace $\mathbf{Y}_{0}$) and hence this is a \textit{soft} intervention on $\mathbf{N}$ that rewrites causal mechanisms while maintaining a causal dependence between parent and child.

Under this new alignment, the high-level interchange intervention $\intinv(\conjMod, \{[\True,\True]\}, \{\{V_1\}\}) = 
\conjMod_{\{V_{1}\}\leftarrow\mathsf{GetVals}_{\{V_1\}}(\conjMod({[\True,\True]}))}$
is aligned with the low-level distributed interchange intervention 
\[
\dintinv(\conjNetwork,\Bigg [ \begin{array}{rr}
      \cos(-\rotdeg) & -\sin(-\rotdeg)\\
      \sin(-\rotdeg) & \phantom{-}\cos (-\rotdeg)
    \end{array}\Bigg ], \{[1,1]\}, \{\{Y_1\}\}) 
\]
and the counterfactual output behavior of $\conjMod$ and $\conjNetwork$ are equal:
\begin{center}   
      \resizebox{0.95\textwidth}{!}{
  \begin{tikzpicture}
    
    \booleanNetworkc{12}{1}{}{red}
    \booleanNetworkccNoOutput{-7.3}{1}{}{}

    \node[fill=red!20, draw, circle, rounded corners=3pt, minimum width=10pt, minimum height=20pt] (Y1) at (4,0.9){$1.0$};
    \node[fill=red!20, draw, circle, rounded corners=3pt, minimum width=10pt, minimum height=20pt] (Y2) at (5,0.9){$1.0$};
    \node[draw, rectangle, rounded corners=3pt, minimum width=60pt, minimum height=32pt] (anchor1) at (4.6,0.9){};
    \node[draw, rectangle, rounded corners=3pt, minimum width=160pt, minimum height=20pt] (anchor2) at (13.25,0.73){};
    \path(anchor2) edge[align, bend right=12pt,->, solid] (anchor1);
    
    \node[fill=yellow!20, draw, circle, rounded corners=3pt, minimum width=10pt, minimum height=20pt] (W2) at (2,0.9){$0.0$};
    \node[fill=yellow!20, draw, circle, rounded corners=3pt, minimum width=10pt, minimum height=20pt] (W1) at (3,0.9){$1.0$};
    \node[draw, rectangle, rounded corners=3pt, minimum width=60pt, minimum height=32pt] (anchor1) at (2.4,0.9){ };
    \node[draw, rectangle, rounded corners=3pt, minimum width=160pt, minimum height=20pt] (anchor2) at (-6.25,0.73){ };
    \path(anchor2) edge[align, bend left=12pt,->, solid] (anchor1);

        \node (rot) at (8,0.6){$    \Bigg [ \begin{array}{rr}
      \cos(-\rotdeg) & -\sin(-\rotdeg)\\
      \sin(-\rotdeg) & \phantom{-}\cos (-\rotdeg)
    \end{array}\Bigg ]$};
    
        \node (rot) at (-1,0.6){$    \Bigg [ \begin{array}{rr}
      \cos(-\rotdeg) & -\sin(-\rotdeg)\\
      \sin(-\rotdeg) & \phantom{-}\cos (-\rotdeg)
    \end{array}\Bigg ]$};

    \node[draw, rectangle, rounded corners=3pt, minimum width=160pt, minimum height=20pt] (anchor2) at (-6.25,2.63){ };
    \path(Y1.north) edge[align, ->, bend right=12pt] (anchor2);
    \path(W1.north) edge[align, ->, bend right=12pt] (anchor2);
        \node (rot2) at (0,3.2){$    \Bigg [ \begin{array}{rr}
      \cos(\rotdeg) & -\sin(\rotdeg)\\
      \sin(\rotdeg) & \phantom{-}\cos (\rotdeg)
    \end{array}\Bigg ]$};
    
    \booleanNetworkbNoInput{-7.25}{2.1}{p}{q}
    
  \end{tikzpicture}
}
\end{center}

In what follows we will assume that $\mathbf{X}$ are already vector spaces (which is true for neural nets) and the functions $\mathbf{R}$ are rotation operators. 
In this case, the subspaces $\mathbf{Y}_j$ can be identified without loss of generality with those spanned by the first $|\mathbf{Y}_0|$ basis vectors for $\mathbf{Y}_0$, the next  $|\mathbf{Y}_1|$ basis vectors for $\mathbf{Y}_1$, and so on.
(The following methods would be well-defined for non-linear transformations, as long as they were invertible and differentiable, but efficient implementation becomes harder.)

\subsection{Distributed Alignment Search}

The question then arises of how to find good rotations. As we discussed above, previous causal abstraction analyses of neural networks have performed brute-force search through a discrete space of hand-picked alignments. In \methodname\ (DAS), we find an alignment between one or more high-level variables and disjoint sub-spaces (but not necessarily subsets) of a large neural representation. 
We define a distributed interchange intervention training objective, use differentiable parameterizations for the space of orthogonal matrices (such as provided by PyTorch), and then optimize the objective with stochastic gradient descent. Crucially, the low-level and high-level models are frozen during learning so we are only changing the alignment.

In the following definition we assume that a neural network specifies an output \textit{distribution} for a given input, which can then be pushed forward to a distribution on output values of the high-level model via an alignment function $\tau$.
We may similarly interpret even a deterministic high-level model as defining a (e.g., delta) distribution on output values.
We make use of these distributions, after interchange intervention, to define a differentiable loss for the rotation matrix which aligns intermediate variables.
\begin{definition}{\textnormal{Distributed Interchange Intervention Training Objective}}\label{def:trainobj}
Begin with a low-level neural network $\mathcal{L}$, with low-level input settings $\mathbf{Inputs}_L$, a high-level algorithm $\mathcal{H}$, with high-level output settings $\mathbf{Out}_H$, and an alignment $\tau$ for their input and output variables. Suppose we want to align intermediate high level variables $X_j \in \mathbf{Vars}_{\mathcal{H}}$ with rotated subspaces $\mathbf{Y}_j$ of a neural representation $\mathbf{N} \subset \mathbf{Vars}_{\mathcal{L}}$ with learned rotation matrix $\mathbf{R}^\theta:\mathbf{N} \to \mathbf{Y}$. 

In general, we can define a training objective using any differentiable loss function $\mathsf{Loss}$ that quantifies the distance between two total high-level settings.
\[
\sum_{\mathbf{b},\mathbf{s}_1, \dots, \mathbf{s}_k \in \mathbf{Inputs}_L} \hspace{-10pt} \mathsf{Loss}\biggl(\dintinv(\mathcal{L},\mathbf{R}^{\theta}, \sources, \{\mathbf{Y}_j\}^k_0)(\mathbf{b}), \\
      \intinv(\mathcal{H}, \{\tau(\mathbf{s}_j)\}^k_1,\sites)(\tau(\mathbf{b}))\biggl)
\]
For our experiments, we compute the cross entropy loss $\mathsf{CE}(\cdot,\cdot)$ between the high-level output distribution $\mathbb{P}(\mathbf{out}_H|\mathcal{H}( \tau(\mathbf{b})))$ and the push-forward under $\tau$ of the low-level output distribution $\mathbb{P}^\tau(\mathbf{out}_H|\mathcal{L}(\mathbf{b}))$. The overall objective is:
\begin{multline*}
\sum_{\mathbf{b},\mathbf{s}_1, \dots, \mathbf{s}_k \in \mathbf{Inputs}_L} \hspace{-20pt} \mathsf{CE}\biggl(
      \mathbb{P}(\mathbf{out}_H|\intinv(\mathcal{H}, \{\tau(\mathbf{s}_j)\}^k_1,\sites))(\tau(\mathbf{b})),
      \mathbb{P}^\tau(\mathbf{out}_H|\dintinv(\mathcal{L},\mathbf{R}^{\theta}, \sources, \{\mathbf{Y}_j\}^k_0)(\mathbf{b}))\biggl)
\end{multline*}
                
\end{definition}
While we still have discrete hyperparameters $(\mathbf{N}, |\mathbf{Y}_0|, \dots, |\mathbf{Y}_k|)$---the target population and the dimensionality of the sub-spaces used for each high-level variable---we may use stochastic gradient descent to determine the rotation that minimizes loss, thus yielding the best distributed alignment between $\mathcal{L}$ and $\mathcal{H}$.

\subsection{Approximate Causal Abstraction}

Perfect causal abstraction relationships are unlikely to arise for neural networks trained to solve complex empirical tasks. We use a graded notion of accuracy:
\begin{definition}{\textnormal{Distributed Interchange Intervention Accuracy }}\label{def:intinvacc}
Given low-level and high-level causal models $\mathcal{L}$ and $\mathcal{H}$ with alignment $(\Pi, \tau)$, rotation $\mathbf{R}:\mathbf{N} \to \mathbf{Y}$, and orthogonal decomposition $\{\mathbf{Y}_j\}^k_0$. If we let $\mathbf{Inputs}_L$ be low-level input settings and $\sites$ be high-level intermediate variables the \textbf{interchange intervention accuracy (\intaccmetricabbr)} is as follows
\[\sum_{\mathbf{b},\mathbf{s}_1,\dots, \mathbf{s}_k \in \mathbf{Inputs}_L} \frac{1}{|\mathbf{Inputs}_{L}|^{k+1} } \Big[\tau\big (\dintinv(\mathcal{L}, \rmat^{\theta},  \sources, \{\mathbf{Y}_j\}^k_0)(\mathbf{b})\big) =
      \intinv(\mathcal{H},  \{\tau(\mathbf{s}_j)\}^k_1, \sites)(\tau(\mathbf{b}))\Big]\]
\end{definition}
\intaccmetricabbr\ is the proportion of aligned interchange interventions that have equivalent high-level and low-level effects. In our example with $\mathcal{N}$ and $\mathcal{A}$,  \intaccmetricabbr\ is 100\% and the high-level model is a perfect abstraction of the low-level model (Def. \ref{def:constructive-abstraction}). When \intaccmetricabbr\ is $\alpha<$100\%, we rely on the graded notion of $\alpha$\textit{-on-average} approximate causal abstraction \citep{geigericard}, which  coincides with \intaccmetricabbr.

%% file: figures/EqualityDistInterchange.tex
\begin{figure*}[t]

\centering
                  \resizebox{\textwidth}{!}{

\begin{tikzpicture}[thick,scale=0.6, every node/.style={scale=0.6}]
\def\x{-6.2}
\def\y{-10.5}
\def\z{0}
\def\w{2}
\node (brown0) at (\x,\y,\z) {\begin{tikzpicture}[scale=1.2]
\def\aa{3}
\def\bb{2}
\def\cc{-2}
\def\rx{35}
\def\ry{35}
\def\rz{35}
  \draw [->,thick=1mm, brown,rotate around x=\rx,  rotate around y=\ry, rotate around z=\rz] (0,0,0) -- (\aa,\bb,\cc) node [at end, left]()  {};
  \draw [->,thick=1mm] (0,0,0) -- (\w,0,0) node [at end, right]() {\huge $\mathbf{X}_1$};
  \draw [->,thick=1mm] (0,0,0) -- (0,\w,0) node [at end, left]()  {\huge $\mathbf{X}_2$};
  \draw [->,thick=1mm] (0,0,0) -- (0,0,\w) node [at end, left]()  {\huge $\mathbf{X}_3$};
\end{tikzpicture}};

\def\x{-5.2}
\def\y{-5}
\def\z{0}
\def\w{3}

\node (brown) at (\x,\y,\z) {\begin{tikzpicture}[scale=1.2]
\def\aa{3}
\def\bb{2}
\def\cc{-2}
\def\rx{35}
\def\ry{35}
\def\rz{35}
  \draw [->, brown,rotate around x=\rx,  rotate around y=\ry, rotate around z=\rz] (0,0,0) -- (\aa,\bb,\cc) node [at end, left]()  {};
  \draw [thick=1mm,->, rotate around x=\rx,  rotate around y=\ry, rotate around z=\rz] (0,0,0) -- (0 + \w,0,0) node [at end, right]() {\Huge $\mathbf{Y}_1$};
  \draw [thick=1mm,->, rotate around x=\rx,  rotate around y=\ry, rotate around z=\rz] (0,0,0) -- (0,\w,0) node [at end, left]()  {\Huge $\mathbf{Y}_2$};
  \draw [thick=1mm,->, rotate around x=\rx,  rotate around y=\ry, rotate around z=\rz] (0,0,0) -- (0,0,\w) node [at end, left]()  {\Huge $\mathbf{Y}_3$};
\def\a{3}
\def\b{2.2}
\def\c{1}
  \node[rotate around x=\rx,  rotate around y=\ry, rotate around z=\rz] (start) at (\a,0, 0) {};
  \draw [dashed, green, rotate around x=\rx,  rotate around y=\ry, rotate around z=\rz] (start) -- (0,0,0);
  \draw [dashed, green,rotate around x=\rx,  rotate around y=\ry, rotate around z=\rz] (start) -- (\aa,\bb,\cc);
\def\a{-0.3}
\def\b{2}
\def\c{-1.5}
  \node[rotate around x=\rx,  rotate around y=\ry, rotate around z=\rz] (start) at (0,0, \c) {};
  \draw [dashed, red,rotate around x=\rx,  rotate around y=\ry, rotate around z=\rz] (start) -- (0,0,0);
  \draw [dashed, red,rotate around x=\rx,  rotate around y=\ry, rotate around z=\rz] (start) -- (\aa,\bb,\cc);
\def\a{1}
\def\b{-3}
\def\c{-2}
  \node[rotate around x=\rx,  rotate around y=\ry, rotate around z=\rz] (start) at (0,\b, 0) {};
  \draw [dashed, blue,rotate around x=\rx,  rotate around y=\ry, rotate around z=\rz] (start) -- (0,0,0);
  \draw [dashed, blue,rotate around x=\rx,  rotate around y=\ry, rotate around z=\rz] (start) -- (\aa,\bb,\cc);
\end{tikzpicture}};
  
\def\x{-22}
\def\y{3}
\def\z{0}
\def\w{3}
\node (green) at (\x,\y,\z) {\begin{tikzpicture}[scale=1.2]
\def\a{3}
\def\b{2.2}
\def\c{1}
  \draw [->, green] (0,0,0) -- (\a,\b,\c) node [at end, left]()  {};
\def\rx{35}
\def\ry{35}
\def\rz{35}
  \draw [thick=1mm,->, rotate around x=\rx,  rotate around y=\ry, rotate around z=\rz] (0,0,0) -- (0 + \w,0,0) node [at end, right]() {\Huge $\mathbf{Y}_1$};
  \draw [thick=1mm,->, rotate around x=\rx,  rotate around y=\ry, rotate around z=\rz] (0,0,0) -- (0,\w,0) node [at end, left]()  {\Huge $\mathbf{Y}_2$};
  \draw [thick=1mm,->, rotate around x=\rx,  rotate around y=\ry, rotate around z=\rz] (0,0,0) -- (0,0,\w) node [at end, left]()  {\Huge $\mathbf{Y}_3$};

  \node[rotate around x=\rx,  rotate around y=\ry, rotate around z=\rz] (start) at (\a,0, 0) {};
  \draw [dashed, green] (start) -- (\a,\b,\c);
  \draw [dashed, green] (start) -- (0,0,0);
\end{tikzpicture}};
  
\def\x{-5.2}
\def\y{3}
\def\z{0}
\def\w{3}

\node (red) at (\x,\y,\z) {\begin{tikzpicture}[scale=1.2]  
\def\a{-0.3}
\def\b{2}
\def\c{-1.5}
  \draw [->, red] (0,0,0) -- (\a,\b,\c) node [at end, left]()  {};
\def\rx{35}
\def\ry{35}
\def\rz{35}
  \draw [thick=1mm,->, rotate around x=\rx,  rotate around y=\ry, rotate around z=\rz] (0,0,0) -- (0 + \w,0,0) node [at end, right]() {\Huge $\mathbf{Y}_1$};
  \draw [thick=1mm,->, rotate around x=\rx,  rotate around y=\ry, rotate around z=\rz] (0,0,0) -- (0,\w,0) node [at end, left]()  {\Huge $\mathbf{Y}_2$};
  \draw [thick=1mm,->, rotate around x=\rx,  rotate around y=\ry, rotate around z=\rz] (0,0,0) -- (0,0,\w) node [at end, left]()  {\Huge $\mathbf{Y}_3$};
  \node[rotate around x=\rx,  rotate around y=\ry, rotate around z=\rz] (start) at (0,0, \c) {};
  \draw [dashed, red] (start) -- (\a,\b,\c);
  \draw [dashed, red] (start) -- (0,0,0);
\end{tikzpicture}};
  
\def\x{12.7}
\def\y{3}
\def\z{0}
\def\w{2}

\node (blue) at (\x,\y,\z) {\begin{tikzpicture}[scale=1.2]
\def\a{1}
\def\b{-3}
\def\c{-2}
  \draw [->, blue] (0,0,0) -- (\a,\b,\c) node [at end, left]()  {};
\def\rx{35}
\def\ry{35}
\def\rz{35}
  \draw [thick=1mm,->, rotate around x=\rx,  rotate around y=\ry, rotate around z=\rz] (0,0,0) -- (0 + \w,0,0) node [at end, right]() {\Huge $\mathbf{Y}_1$};
  \draw [thick=1mm,->, rotate around x=\rx,  rotate around y=\ry, rotate around z=\rz] (0,0,0) -- (0,\w,0) node [at end, left]()  {\Huge $\mathbf{Y}_2$};
  \draw [thick=1mm,->, rotate around x=\rx,  rotate around y=\ry, rotate around z=\rz] (0,0,0) -- (0,0,\w) node [at end, left]()  {\Huge $\mathbf{Y}_3$};
  \node[rotate around x=\rx,  rotate around y=\ry, rotate around z=\rz] (start) at (0,\b, 0) {};
  \draw [dashed, blue] (start) -- (\a,\b,\c);
  \draw [dashed, blue] (start) -- (0,0,0);
\end{tikzpicture}};

\def\x{-22}
\def\y{10}
\def\z{0}
\def\w{2}

  \draw [->] (\x,\y,\z) -- (\x + \w,\y,\z) node [at end, right]() {\huge $\mathbf{X}_1$};
  \draw [->] (\x,\y,\z) -- (\x,\y+\w,\z) node [at end, left]()  {\huge $\mathbf{X}_2$};
  \draw [->] (\x,\y,\z) -- (\x,\y,\z + \w) node [at end, left]()  {\huge $\mathbf{X}_3$};
  
\def\a{3}
\def\b{2.2}
\def\c{1}
  \draw [->, green] (\x,\y,\z) -- (\x+\a,\y + \b,\z + \c) node [at end, left]()  {};
  
\def\x{-4}
\def\y{10}
\def\z{0}
\def\w{2}

  \draw [->] (\x,\y,\z) -- (\x + \w,\y,\z) node [at end, right]() {\huge $\mathbf{X}_1$};
  \draw [->] (\x,\y,\z) -- (\x,\y+\w,\z) node [at end, left]()  {\huge $\mathbf{X}_2$};
  \draw [->] (\x,\y,\z) -- (\x,\y,\z + \w) node [at end, left]()  {\huge $\mathbf{X}_3$};
  
\def\a{0.3}
\def\b{2}
\def\c{-0.5}
  \draw [->, red] (\x,\y,\z) -- (\x+\a,\y + \b,\z + \c) node [at end, left]()  {};
  
\def\x{13}
\def\y{10}
\def\z{0}
\def\w{2}

  \draw [->] (\x,\y,\z) -- (\x + \w,\y,\z) node [at end, right]() {\huge $\mathbf{X}_1$};
  \draw [->] (\x,\y,\z) -- (\x,\y+\w,\z) node [at end, left]()  {\huge $\mathbf{X}_2$};
  \draw [->] (\x,\y,\z) -- (\x,\y,\z + \w) node [at end, left]()  {\huge $\mathbf{X}_3$};
  
\def\a{1}
\def\b{-3}
\def\c{-2}
  \draw [->, blue] (\x,\y,\z) -- (\x+\a,\y + \b,\z + \c) node [at end, left]()  {};

\def\scalex{3.5}
\def\scaley{1.95}
\def\opac{0.2}
\Large
\def\startx{-17}
\def\starty{7}

\node (label) at (1.5 * \scalex+ \startx,3.25 * \scaley+ \starty) {};

\node[fill=green!20, draw, rectangle, rounded corners=3pt, minimum width=40pt, minimum height=20pt](out) at (1.5 * \scalex+ \startx,2.75 * \scaley+ \starty) {};

\node[fill=green!20, draw, rectangle, rounded corners=3pt, minimum width=40pt, minimum height=20pt] (BERT00) at (0 * \scalex+ \startx,0 * \scaley + \starty) {};
\node[fill=green!20, draw, rectangle, rounded corners=3pt, minimum width=40pt, minimum height=20pt] (BERT01) at (1 * \scalex + \startx,0 * \scaley+ \starty) {};

\node[fill=green!20, draw, rectangle, rounded corners=3pt, minimum width=40pt, minimum height=20pt] (BERT02) at (2 * \scalex+ \startx,0 * \scaley+ \starty) {};
\node[fill=green!20, draw, rectangle, rounded corners=3pt, minimum width=40pt, minimum height=20pt] (BERT03) at (3 * \scalex + \startx,0 * \scaley+ \starty) {};

\node[fill=green!20, draw, rectangle, rounded corners=3pt, minimum width=40pt, minimum height=20pt] (BERT10) at (0 * \scalex+ \startx,1 * \scaley+ \starty){$\mathbf{X}_1$};

\node[fill=green!20, draw, rectangle, rounded corners=3pt, minimum width=40pt, minimum height=20pt] (BERT11) at (1 * \scalex+ \startx,1 * \scaley+ \starty){$\mathbf{X}_2$};

\node[fill=green!20, draw, rectangle, rounded corners=3pt, minimum width=40pt, minimum height=20pt] (BERT12) at (2 * \scalex+ \startx,1 * \scaley+ \starty){$\mathbf{X}_3$};

\node[fill=green!20, draw, rectangle, rounded corners=3pt, minimum width=40pt, minimum height=20pt] (BERT13) at (3 * \scalex+ \startx,1 * \scaley+ \starty){};

\node[fill=green!20, draw, rectangle, rounded corners=3pt, minimum width=40pt, minimum height=20pt] (BERT20) at (0 * \scalex+ \startx,2 * \scaley+ \starty){};

\node[fill=green!20, draw, rectangle, rounded corners=3pt, minimum width=40pt, minimum height=20pt] (BERT21) at (1 * \scalex+ \startx,2 * \scaley+ \starty){};

\node[fill=green!20, draw, rectangle, rounded corners=3pt, minimum width=40pt, minimum height=20pt] (BERT22) at (2 * \scalex+ \startx,2 * \scaley+ \starty){};

\node[fill=green!20, draw, rectangle, rounded corners=3pt, minimum width=40pt, minimum height=20pt] (BERT23) at (3 * \scalex+ \startx,2 * \scaley+ \starty){};

\foreach \col in {0,1,2,3}{
\foreach \coll in {0,1,2,3}{
\draw[opacity=0.3, ->] (BERT0\col.north) to (BERT1\coll.south);
\draw[opacity=0.3, ->] (BERT1\col.north) to (BERT2\coll.south);
}
}

\foreach \col in {0,1,2,3}{
\draw[opacity=0.3, ->] (BERT2\col.north) to (out.south);
}

\node[draw, rectangle,dashed, minimum width=300,minimum height=36pt] (anchor1) at (1 * \scalex+ \startx,1 * \scaley+ \starty) {};

\def\startx{0}
\def\starty{7}

\node (label) at (1.5 * \scalex+ \startx,3.25 * \scaley+ \starty) {};

\node[fill=red!20, draw, rectangle, rounded corners=3pt, minimum width=40pt, minimum height=20pt](out) at (1.5 * \scalex+ \startx,2.75 * \scaley+ \starty) {};

\node[fill=red!20, draw, rectangle, rounded corners=3pt, minimum width=40pt, minimum height=20pt] (BERT00) at (0 * \scalex+ \startx,0 * \scaley + \starty) {};
\node[fill=red!20, draw, rectangle, rounded corners=3pt, minimum width=40pt, minimum height=20pt] (BERT01) at (1 * \scalex + \startx,0 * \scaley+ \starty) {};

\node[fill=red!20, draw, rectangle, rounded corners=3pt, minimum width=40pt, minimum height=20pt] (BERT02) at (2 * \scalex+ \startx,0 * \scaley+ \starty) {};
\node[fill=red!20, draw, rectangle, rounded corners=3pt, minimum width=40pt, minimum height=20pt] (BERT03) at (3 * \scalex + \startx,0 * \scaley+ \starty) {};

\node[fill=red!20, draw, rectangle, rounded corners=3pt, minimum width=40pt, minimum height=20pt] (BERT10) at (0 * \scalex+ \startx,1 * \scaley+ \starty){$\mathbf{X}_1$};

\node[fill=red!20, draw, rectangle, rounded corners=3pt, minimum width=40pt, minimum height=20pt] (BERT11) at (1 * \scalex+ \startx,1 * \scaley+ \starty){$\mathbf{X}_2$};

\node[fill=red!20, draw, rectangle, rounded corners=3pt, minimum width=40pt, minimum height=20pt] (BERT12) at (2 * \scalex+ \startx,1 * \scaley+ \starty){$\mathbf{X}_3$};

\node[fill=red!20, draw, rectangle, rounded corners=3pt, minimum width=40pt, minimum height=20pt] (BERT13) at (3 * \scalex+ \startx,1 * \scaley+ \starty){};

\node[fill=red!20, draw, rectangle, rounded corners=3pt, minimum width=40pt, minimum height=20pt] (BERT20) at (0 * \scalex+ \startx,2 * \scaley+ \starty){};

\node[fill=red!20, draw, rectangle, rounded corners=3pt, minimum width=40pt, minimum height=20pt] (BERT21) at (1 * \scalex+ \startx,2 * \scaley+ \starty){};

\node[fill=red!20, draw, rectangle, rounded corners=3pt, minimum width=40pt, minimum height=20pt] (BERT22) at (2 * \scalex+ \startx,2 * \scaley+ \starty){};

\node[fill=red!20, draw, rectangle, rounded corners=3pt, minimum width=40pt, minimum height=20pt] (BERT23) at (3 * \scalex+ \startx,2 * \scaley+ \starty){};

\foreach \col in {0,1,2,3}{
\foreach \coll in {0,1,2,3}{
\draw[opacity=0.3, ->] (BERT0\col.north) to (BERT1\coll.south);
\draw[opacity=0.3, ->] (BERT1\col.north) to (BERT2\coll.south);
}
}

\foreach \col in {0,1,2,3}{
\draw[opacity=0.3, ->] (BERT2\col.north) to (out.south);
}

\node[draw,  rectangle,dashed, minimum width=300pt,minimum height=36pt] (anchor1b) at (1 * \scalex+ \startx,1 * \scaley+ \starty) {};

\def\opac{1.0}

\def\scalex{3.5}
\def\scaley{1.95}
\def\opac{0.2}
\Large

\def\startx{17}

\node (label) at (1.5 * \scalex+ \startx,3.25 * \scaley+ \starty) {};

\node[fill=blue!20, draw, rectangle, rounded corners=3pt, minimum width=40pt, minimum height=20pt](out) at (1.5 * \scalex+ \startx,2.75 * \scaley+ \starty) {};

\node[fill=blue!20, draw, rectangle, rounded corners=3pt, minimum width=40pt, minimum height=20pt] (BERT00) at (0 * \scalex+ \startx,0 * \scaley+ \starty) {};
\node[fill=blue!20, draw, rectangle, rounded corners=3pt, minimum width=40pt, minimum height=20pt] (BERT01) at (1 * \scalex + \startx,0 * \scaley+ \starty) {};

\node[fill=blue!20, draw, rectangle, rounded corners=3pt, minimum width=40pt, minimum height=20pt] (BERT02) at (2 * \scalex+ \startx,0 * \scaley+ \starty) {};
\node[fill=blue!20, draw, rectangle, rounded corners=3pt, minimum width=40pt, minimum height=20pt] (BERT03) at (3 * \scalex + \startx,0 * \scaley+ \starty) {};

\node[fill=blue!20, draw, rectangle, rounded corners=3pt, minimum width=40pt, minimum height=20pt] (BERT10) at (0 * \scalex+ \startx,1 * \scaley+ \starty){$\mathbf{X}_1$};

\node[fill=blue!20, draw, rectangle, rounded corners=3pt, minimum width=40pt, minimum height=20pt] (BERT11) at (1 * \scalex+ \startx,1 * \scaley+ \starty){$\mathbf{X}_2$};

\node[fill=blue!20, draw, rectangle, rounded corners=3pt, minimum width=40pt, minimum height=20pt] (BERT12) at (2 * \scalex+ \startx,1 * \scaley+ \starty){$\mathbf{X}_3$};

\node[fill=blue!20, draw, rectangle, rounded corners=3pt, minimum width=40pt, minimum height=20pt] (BERT13) at (3 * \scalex+ \startx,1 * \scaley+ \starty){};

\node[fill=blue!20, draw, rectangle, rounded corners=3pt, minimum width=40pt, minimum height=20pt] (BERT20) at (0 * \scalex+ \startx,2 * \scaley+ \starty){};

\node[fill=blue!20, draw, rectangle, rounded corners=3pt, minimum width=40pt, minimum height=20pt] (BERT21) at (1 * \scalex+ \startx,2 * \scaley+ \starty){};

\node[fill=blue!20, draw, rectangle, rounded corners=3pt, minimum width=40pt, minimum height=20pt] (BERT22) at (2 * \scalex+ \startx,2 * \scaley+ \starty){};

\node[fill=blue!20, draw, rectangle, rounded corners=3pt, minimum width=40pt, minimum height=20pt] (BERT23) at (3 * \scalex+ \startx,2 * \scaley+ \starty){};

\foreach \col in {0,1,2,3}{
\foreach \coll in {0,1,2,3}{
\draw[opacity=0.3, ->] (BERT0\col.north) to (BERT1\coll.south);
\draw[opacity=0.3, ->] (BERT1\col.north) to (BERT2\coll.south);
}
}

\foreach \col in {0,1,2,3}{
\draw[opacity=0.3, ->] (BERT2\col.north) to (out.south);
}
\node[draw, rectangle,dashed, minimum width=300pt,minimum height=36pt] (anchor1c) at (1 * \scalex+ \startx,1 * \scaley+ \starty) {};

\def\scalex{3.5}
\def\opac{0.2}
\Large
\def\startx{-17}
\def\starty{-1}

\node (label) at (1.5 * \scalex+ \startx,3.25 * \scaley+ \starty) {};

\node[fill=green!20, draw, circle, rounded corners=3pt, minimum width=40pt, minimum height=20pt] (Y1a) at (0 * \scalex+ \startx,1 * \scaley+ \starty){$\mathbf{Y}_1$};

\node[fill=green!20, draw, circle, rounded corners=3pt, minimum width=40pt, minimum height=20pt] (Y2a) at (1 * \scalex+ \startx,1 * \scaley+ \starty){$\mathbf{Y}_2$};
\node[draw, rectangle, dashed, rounded corners=3pt, minimum width=300pt, minimum height=64pt] (anchor2) at (1 * \scalex+ \startx,1 * \scaley+ \starty){};

\node[fill=green!20, draw, circle, rounded corners=3pt, minimum width=40pt, minimum height=20pt] (Y3a) at (2 * \scalex+ \startx,1 * \scaley+ \starty){$\mathbf{Y}_3$};

\draw[->, line width=1mm] (anchor1) -- (anchor2) node[midway, above right]{{\Large $\mathbf{R}$}};

\def\startx{0}

\def\starty{-1}

\node (label) at (1.5 * \scalex+ \startx,3.25 * \scaley+ \starty) {};

\node[fill=red!20, draw, circle, rounded corners=3pt, minimum width=40pt, minimum height=20pt] (Y1b) at (0 * \scalex+ \startx,1 * \scaley+ \starty){$\mathbf{Y}_1$};

\node[fill=red!20, draw, circle, rounded corners=3pt, minimum width=40pt, minimum height=20pt] (Y2b) at (1 * \scalex+ \startx,1 * \scaley+ \starty){$\mathbf{Y}_2$};
\node[draw, rectangle, dashed, rounded corners=3pt, minimum width=300pt, minimum height=64pt] (anchor2b) at (1 * \scalex+ \startx,1 * \scaley+ \starty){};

\node[fill=red!20, draw, circle, rounded corners=3pt, minimum width=40pt, minimum height=20pt] (Y2d) at (2 * \scalex+ \startx,1 * \scaley+ \starty){$\mathbf{Y}_3$};

\draw[->, line width=1mm] (anchor1b) -- (anchor2b) node[midway, above right]{{\Large $\mathbf{R}$}};

\def\starty{-7}

\node[fill=green!20, draw, circle, rounded corners=3pt, minimum width=40pt, minimum height=20pt] (Y1b) at (0 * \scalex+ \startx,1 * \scaley+ \starty){$\mathbf{Y}_1$};

\node[fill=blue!20, draw, circle, rounded corners=3pt, minimum width=40pt, minimum height=20pt] (Y2b) at (1 * \scalex+ \startx,1 * \scaley+ \starty){$\mathbf{Y}_2$};
\node[draw, rectangle, dashed, rounded corners=3pt, minimum width=300pt, minimum height=64pt] (anchor3b) at (1 * \scalex+ \startx,1 * \scaley+ \starty){};

\node[fill=red!20, draw, circle, rounded corners=3pt, minimum width=40pt, minimum height=20pt] (Y3b) at (2 * \scalex+ \startx,1 * \scaley+ \starty){$\mathbf{Y}_3$};

\def\starty{8.54}
\def\opac{1.0}

\def\scalex{3.5}
\def\scaley{1.95}
\def\opac{0.2}
\Large

\def\startx{17}
\def\starty{-1}

\node[fill=blue!20, draw, circle, rounded corners=3pt, minimum width=40pt, minimum height=20pt] (Y1c) at (0 * \scalex+ \startx,1 * \scaley+ \starty){$\mathbf{Y}_1$};

\node[fill=blue!20, draw, circle, rounded corners=3pt, minimum width=40pt, minimum height=20pt] (Y2c) at (1 * \scalex+ \startx,1 * \scaley+ \starty){$\mathbf{Y}_2$};

\node[draw, rectangle, dashed, rounded corners=3pt, minimum width=300pt, minimum height=64pt] (anchor2c) at (1 * \scalex+ \startx,1 * \scaley+ \starty){};

\node[fill=blue!20, draw, circle, rounded corners=3pt, minimum width=40pt, minimum height=20pt] (Y3c) at (2 * \scalex+ \startx,1 * \scaley+ \starty){$\mathbf{Y}_3$};

\draw[->, line width=1mm] (anchor1c) -- (anchor2c) node[midway, above right]{{\Large $\mathbf{R}$}};

\draw[opacity=0.8, dashed, ->, thick] (Y1a) to[in=90, out=-35] (Y1b);

\draw[opacity=0.8, dashed, ->, thick] (Y2c) to[in=55, out=-100] (Y2b);

\draw[opacity=0.8, dashed, ->, thick] (Y2d) to[in=70, out=-60] (Y3b);

\def\startx{0}

\def\starty{-13.5}

\node (label) at (1.5 * \scalex+ \startx,3.25 * \scaley+ \starty) {};

\node[fill=brown!20, draw, rectangle, rounded corners=3pt, minimum width=40pt, minimum height=20pt](out) at (1.5 * \scalex+ \startx,2.75 * \scaley+ \starty) {};

\node[draw,  rectangle,dashed, minimum width=300pt,minimum height=36pt] (anchor2) at (1 * \scalex+ \startx,1 * \scaley+ \starty) {};

\node[fill=red!20, draw, rectangle, rounded corners=3pt, minimum width=40pt, minimum height=20pt] (BERT00) at (0 * \scalex+ \startx,0 * \scaley+ \starty) {};
\node[fill=red!20, draw, rectangle, rounded corners=3pt, minimum width=40pt, minimum height=20pt] (BERT01) at (1 * \scalex + \startx,0 * \scaley+ \starty) {};

\node[fill=red!20, draw, rectangle, rounded corners=3pt, minimum width=40pt, minimum height=20pt] (BERT02) at (2 * \scalex+ \startx,0 * \scaley+ \starty) {};
\node[fill=red!20, draw, rectangle, rounded corners=3pt, minimum width=40pt, minimum height=20pt] (BERT03) at (3 * \scalex + \startx,0 * \scaley+ \starty) {};

\node[fill=brown!20, draw, rectangle, rounded corners=3pt, minimum width=40pt, minimum height=20pt] (BERT10) at (0 * \scalex+ \startx,1 * \scaley+ \starty){$\mathbf{X}_1$};

\node[fill=brown!20, draw, rectangle, rounded corners=3pt, minimum width=40pt, minimum height=20pt] (BERT11) at (1 * \scalex+ \startx,1 * \scaley+ \starty){$\mathbf{X}_2$};

\node[fill=brown!20, draw, rectangle, rounded corners=3pt, minimum width=40pt, minimum height=20pt] (BERT12) at (2 * \scalex+ \startx,1 * \scaley+ \starty){$\mathbf{X}_3$};

\node[fill=red!20, draw, rectangle, rounded corners=3pt, minimum width=40pt, minimum height=20pt] (BERT13) at (3 * \scalex+ \startx,1 * \scaley+ \starty){};

\node[fill=brown!20, draw, rectangle, rounded corners=3pt, minimum width=40pt, minimum height=20pt] (BERT20) at (0 * \scalex+ \startx,2 * \scaley+ \starty){};

\node[fill=brown!20, draw, rectangle, rounded corners=3pt, minimum width=40pt, minimum height=20pt] (BERT21) at (1 * \scalex+ \startx,2 * \scaley+ \starty){ };

\node[fill=brown!20, draw, rectangle, rounded corners=3pt, minimum width=40pt, minimum height=20pt] (BERT22) at (2 * \scalex+ \startx,2 * \scaley+ \starty){};

\node[fill=brown!20, draw, rectangle, rounded corners=3pt, minimum width=40pt, minimum height=20pt] (BERT23) at (3 * \scalex+ \startx,2 * \scaley+ \starty){};

\draw[->, line width=1mm] (anchor3b) -- (anchor2) node[midway, above left]{{\Large $\mathbf{R}^{-1}$}};
\foreach \col in {0,1,2,3}{
\foreach \coll in {0,1,2,3}{
\draw[opacity=0.3, ->] (BERT0\col.north) to (BERT1\coll.south);
\draw[opacity=0.3, ->] (BERT1\col.north) to (BERT2\coll.south);
}
}

\foreach \col in {0,1,2,3}{
\draw[opacity=0.3, ->] (BERT2\col.north) to (out.south);
}

\end{tikzpicture}
}
\caption{A generic multi-source distributed interchange intervention. 
The base input and two source inputs create three total settings of a model. The top left (green) and right (blue) total model settings are determined by two source inputs and the middle total model setting (red) is determined by the base input. Three hidden units from each total setting are rotated with an orthogonal matrix $\mathbf{R}:\mathbf{X}\to\mathbf{Y}$. Then we intervene on the rotated representation for the base input and fix two dimensions to be the value they take on for each source input, respectively. Then we unrotate the representation with $\mathbf{R}^{-1}$ and compute a counterfactual total model setting for the base input. In DAS, the orthogonal matrix is found with gradient descent using a high-level causal model to guide the search process.
}
\label{fig:equalityinterchange-maintext}
\end{figure*}

%% file: tables/eq-results.tex
\begin{table*}[t]
    \centering
    \resizebox{1.0\linewidth}{!}{%
    \begin{tabular}{cc  ccc  ccc  ccc c}
    & & \multicolumn{3}{c}{Both Equality Relations} & \multicolumn{3}{c}{Left Equality Relation} & \multicolumn{3}{c}{Identity of First Argument} & Identity Subspace of \\
    & & \multicolumn{3}{c}{} & \multicolumn{3}{c}{} & \multicolumn{3}{c}{} & Left Equality \\
    & & \multicolumn{3}{c}{\input{graphs/1234}} & \multicolumn{3}{c}{\input{graphs/12}}& \multicolumn{3}{c}{\input{graphs/1}} & \multicolumn{1}{c}{\input{graphs/subspace_12_1}}\\
    Hidden size & Intervention size & \textbf{Layer 1} & \textbf{Layer 2} & \textbf{Layer 3} & \textbf{Layer 1} & \textbf{Layer 2} & \textbf{Layer 3} & \textbf{Layer 1} & \textbf{Layer 2} & \textbf{Layer 3} & \textbf{Layer 1} \\
    \midrule
    $|\mathbf{N}| =16$ & $  1$  &     0.88 &     0.51 &      0.50 &     0.85 &     0.54 &      0.50 &     0.51 &     0.52 &     0.50  & 0.51 \\
    $|\mathbf{N}| =16$ & $  2$  &     0.97 &     0.54 &      0.50 &     0.85 &     0.55 &      0.50 &     0.50 &     0.52 &     0.51  & 0.50 \\
    $|\mathbf{N}| =16$ & $  8$  &     1.00 &     0.57 &      0.50 &     0.90 &     0.56 &      0.50 &     0.52 &     0.53 &     0.51  & 0.51 \\[0.3cm]
    $|\mathbf{N}| =32$ & $  2$  &    0.93 &    0.63 &    0.49 &    0.92 &    0.65 &    0.50 &    0.52 &    0.55 &    0.52  & 0.50 \\
    $|\mathbf{N}| =32$ & $  4$  &    0.97 &    0.63 &    0.49 &    0.94 &    0.65 &    0.50 &    0.51 &    0.55 &    0.52  & 0.51 \\
    $|\mathbf{N}| =32$ & $  16$ &    0.99 &    0.67 &    0.53 &    0.99 &    0.65 &    0.50 &    0.49 &    0.55 &    0.52  & 0.51 \\
    \midrule
     \multicolumn{2}{c}{Brute-Force Search} &   0.60 & 0.56 & 0.52 & 0.64 & 0.64 & 0.57 & 0.50 & 0.51 & 0.54 & - \\
     \multicolumn{2}{c}{Localist Alignment} & 0.73 & 0.56 & 0.48 & 0.60 & 0.50 & 0.49 & 0.46 & 0.47 & 0.48 & - \\
    \bottomrule
    \end{tabular}}
    \caption{Hierarchical equality alignment learning results. The table can be read as follows: \textbf{Layer 1}, \textbf{Layer 2}, and \textbf{Layer 3} indicate which layer of neurons is targeted, $|\mathbf{N}|$ is the number of neurons in a layer, $k$ is the number of neurons aligned with each intermediate variable (\textcolor{red}{red}) where our subspace model occupies $\frac{k}{2}$ with rounding up to the closest integer, and the values in each cell are interchange intervention accuracies for the learned alignment on training data. We report the best results from three runs with distinct random seeds for training the rotation matrix (the same frozen low-level model is used for each seed).}
    \label{tab:equalityresults}
\end{table*}

%% file: graphs/1234.tex
\resizebox{50pt}{!}{
\begin{tikzpicture}[scale=0.6, every node/.style={scale=0.6}]
\def\scalex{2.1}
\def\scaley{2}
\def\startx{0}
\def\starty{-0.43}
\def\opac{1.0}
\def\starty{0}

\node[ draw, circle, minimum size=20pt] (A) at (0+ \startx,0+ \starty) {};

\node[ draw, circle, minimum size=20pt] (B) at (1*\scalex+ \startx,0+ \starty) {};

\node[ draw, circle, minimum size=20pt] (C) at (2*\scalex+ \startx,0+ \starty) {};

\node[ draw, circle, minimum size=20pt] (D) at (3*\scalex+ \startx,0+ \starty) {};

\node[draw, ellipse, minimum size=20pt, fill=red] (E) at (0.5*\scalex+ \startx,0.5*\scaley+ \starty) {};

\node[draw, ellipse, minimum size=20pt, fill=red] (F) at (2.5*\scalex + \startx,0.5*\scaley+ \starty) {};

\node[ draw, ellipse, minimum size=20pt] (G) at (1.5*\scalex + \startx,1*\scaley+ \starty) {};

\draw [->,opacity=\opac] (A) -- (E);
\draw [->,opacity=\opac] (B) -- (E);
\draw [->,opacity=\opac] (C) -- (F);
\draw [->,opacity=\opac] (D) -- (F);
\draw [->,opacity=\opac] (E) -- (G);
\draw [->,opacity=\opac] (F) -- (G);
\end{tikzpicture}
}

%% file: graphs/12.tex
\resizebox{50pt}{!}{
\begin{tikzpicture}[scale=0.6, every node/.style={scale=0.6}]
\def\scalex{2.1}
\def\scaley{2}
\def\startx{0}
\def\starty{-0.43}
\def\opac{1.0}
\def\starty{0}

\node[ draw, circle, minimum size=20pt] (A) at (0+ \startx,0+ \starty) {};

\node[ draw, circle, minimum size=20pt] (B) at (1*\scalex+ \startx,0+ \starty) {};

\node[ draw, circle, minimum size=20pt] (C) at (2*\scalex+ \startx,0+ \starty) {};

\node[ draw, circle, minimum size=20pt] (D) at (3*\scalex+ \startx,0+ \starty) {};

\node[draw, ellipse, minimum size=20pt, fill=red] (E) at (0.5*\scalex+ \startx,0.5*\scaley+ \starty) {};

\node[ draw, ellipse, minimum size=20pt] (G) at (1.5*\scalex + \startx,1*\scaley+ \starty) {};

\draw [->,opacity=\opac] (A) -- (E);
\draw [->,opacity=\opac] (B) -- (E);
\draw [->,opacity=\opac] (C) -- (G);
\draw [->,opacity=\opac] (D) -- (G);
\draw [->,opacity=\opac] (E) -- (G);
\end{tikzpicture}
}

%% file: graphs/1.tex
\resizebox{50pt}{!}{
\begin{tikzpicture}[scale=0.6, every node/.style={scale=0.6}]
\def\scalex{2.1}
\def\scaley{2}
\def\startx{0}
\def\starty{-0.43}
\def\opac{1.0}
\def\starty{0}

\node[ draw, circle, minimum size=20pt] (A) at (0+ \startx,0+ \starty) {};

\node[ draw, circle, minimum size=20pt] (B) at (1*\scalex+ \startx,0+ \starty) {};

\node[ draw, circle, minimum size=20pt] (C) at (2*\scalex+ \startx,0+ \starty) {};

\node[ draw, circle, minimum size=20pt] (D) at (3*\scalex+ \startx,0+ \starty) {};

\node[draw, ellipse, minimum size=20pt, fill=red] (E) at (0.5*\scalex+ \startx,0.5*\scaley+ \starty) {};

\node[ draw, ellipse, minimum size=20pt] (G) at (1.5*\scalex + \startx,1*\scaley+ \starty) {};

\draw [->,opacity=\opac] (A) -- (E);
\draw [->,opacity=\opac] (B) -- (G);
\draw [->,opacity=\opac] (C) -- (G);
\draw [->,opacity=\opac] (D) -- (G);
\draw [->,opacity=\opac] (E) -- (G);
\end{tikzpicture}

}

%% file: graphs/subspace_12_1.tex
\resizebox{50pt}{!}{
\begin{tikzpicture}[scale=0.6, every node/.style={scale=0.6}]
\def\scalex{2.1}
\def\scaley{2}
\def\startx{0}
\def\starty{-0.43}
\def\opac{1.0}
\def\starty{0}

\node[ draw, circle, minimum size=20pt] (A) at (0+ \startx,0+ \starty) {};

\node[ draw, circle, minimum size=20pt] (B) at (1*\scalex+ \startx,0+ \starty) {};

\node[ draw, circle, minimum size=20pt] (C) at (2*\scalex+ \startx,0+ \starty) {};

\node[ draw, circle, minimum size=20pt] (D) at (3*\scalex+ \startx,0+ \starty) {};

\node[draw, ellipse, minimum size=20pt] (E) at (0.5*\scalex+ \startx,0.3*\scaley+ \starty) {};

\node[draw, ellipse, minimum size=20pt, fill=red] (F) at (0.25*\scalex+ \startx,0.8*\scaley+ \starty) {};

\node[draw, ellipse, minimum size=20pt] (G) at (0.75*\scalex+ \startx,0.8*\scaley+ \starty) {};

\node[draw, ellipse, minimum size=20pt] (H) at (0.5*\scalex+ \startx,1.3*\scaley+ \starty) {};

\node[ draw, ellipse, minimum size=20pt] (I) at (1.5*\scalex + \startx,1.8*\scaley+ \starty) {};

\draw [->,opacity=\opac] (A) -- (E);
\draw [->,opacity=\opac] (B) -- (E);

\draw [->,opacity=\opac] (E) -- (F);
\draw [->,opacity=\opac] (E) -- (G);

\draw [->,opacity=\opac] (F) -- (H);
\draw [->,opacity=\opac] (G) -- (H);

\draw [->,opacity=\opac] (H) -- (I);

\draw [->,opacity=\opac] (C) -- (I);
\draw [->,opacity=\opac] (D) -- (I);

\end{tikzpicture}

}

%% file: graphs/1234-large.tex
\begin{tikzpicture}[scale=0.6, every node/.style={scale=0.6}]
\def\scalex{2.1}
\def\scaley{2}
\def\startx{0}
\def\starty{-0.43}
\def\opac{1.0}
\def\starty{0}

\node[ draw, circle, minimum size=20pt] (A) at (0+ \startx,0+ \starty) {$w$};

\node[ draw, circle, minimum size=20pt] (B) at (1*\scalex+ \startx,0+ \starty) {$x$};

\node[ draw, circle, minimum size=20pt] (C) at (2*\scalex+ \startx,0+ \starty) {$y$};

\node[ draw, circle, minimum size=20pt] (D) at (3*\scalex+ \startx,0+ \starty) {$z$};

\node[draw, ellipse, minimum size=20pt, fill=red!60] (E) at (0.5*\scalex+ \startx,0.5*\scaley+ \starty) {$V_{1} \gets (w = x)$};

\node[draw, ellipse, minimum size=20pt, fill=red!60] (F) at (2.5*\scalex + \startx,0.5*\scaley+ \starty) {$V_{2} \gets (y = z)$};

\node[ draw, ellipse, minimum size=20pt] (G) at (1.5*\scalex + \startx,1*\scaley+ \starty) {$V_{3} \gets (V_{1} = V_{2})$};

\draw [->,opacity=\opac] (A) -- (E);
\draw [->,opacity=\opac] (B) -- (E);
\draw [->,opacity=\opac] (C) -- (F);
\draw [->,opacity=\opac] (D) -- (F);
\draw [->,opacity=\opac] (E) -- (G);
\draw [->,opacity=\opac] (F) -- (G);
\end{tikzpicture}

%% file: tables/nli-results.tex
\begin{table}[tp]
    \centering
    \resizebox{1.0\linewidth}{!}{%
    \begin{tabular}{cc  ccc ccc  ccc c}
    & & \multicolumn{3}{c}{Negation and} & \multicolumn{3}{c}{Lexical Entailment} & \multicolumn{3}{c}{Identity of Lexeme} & Lexeme Subspace of \\
    & & \multicolumn{3}{c}{Lexical Entailment} & \multicolumn{3}{c}{} & \multicolumn{3}{c}{} & Lexical Entailment \\
    & & \multicolumn{3}{c}{\input{graphs/neghyp.tex}} & \multicolumn{3}{c}{\input{graphs/neghyp_hyp.tex}}& \multicolumn{3}{c}{\input{graphs/wordid_right.tex}} & \multicolumn{1}{c}{\input{graphs/subspace_wordid.tex}} \\
    Hidden size & Intervention size &\textbf{Layer 7} & \textbf{Layer 9} & \textbf{Layer 11} & \textbf{Layer 7} & \textbf{Layer 9} & \textbf{Layer 11} & \textbf{Layer 7} & \textbf{Layer 9} & \textbf{Layer 11} & \textbf{Layer 9} \\
    \midrule
    $|\mathbf{N}| =768$ & $  64$  &    0.65 &    0.96 &     0.91 &    0.88 &    1.00 &     0.97 &    0.88 &    0.94 &     0.93 & 0.97  \\
    $|\mathbf{N}| =768$ & $  128$  &    0.65 &    0.99 &     0.92 &    0.88 &    1.00 &     0.99 &    0.89 &    0.93 &     0.92 & 0.97 \\
    $|\mathbf{N}| =768$ & $  256$ &    0.67 &    1.00 &     0.86 &    0.91 &    1.00 &     1.00 &    0.88 &    0.96 &     0.88  & 0.98 \\
    \midrule
     \multicolumn{2}{c}{Brute-Force Search} & 0.60 & 0.56 & 0.52 & 0.64 & 0.64 & 0.57 & 0.50 & 0.51 & 0.54 & - \\
     \multicolumn{2}{c}{Localist Alignment} & 0.51 & 0.51 & 0.51 & 0.47 & 0.47 & 0.47 & 0.50 & 0.50 & 0.50 & - \\
    \bottomrule
    \end{tabular}}
    \caption{Monotonicity NLI results. The table can be read as follows: \textbf{Layer 7}, \textbf{Layer 9}, and \textbf{Layer 11} indicate which layer of neurons is targeted, $|\mathbf{N}|$ is the number of neurons in a layer, $k$ is the number of neurons aligned with each intermediate variable (\textcolor{red}{red}) where our subspace model occupies $\frac{k}{2}$, and the values in each cell are interchange intervention accuracies for the learned alignment on training data. We report the best results from three runs with distinct random seeds.}
    \label{tab:monliresults}
\end{table}

%% file: graphs/neghyp.tex
\resizebox{50pt}{!}{
\begin{tikzpicture}[scale=0.6, every node/.style={scale=0.6}]
\def\scalex{2.1}
\def\scaley{2}
\def\startx{0}
\def\starty{-0.43}
\def\opac{1.0}
\def\starty{0}

\node[ draw, circle, minimum size=20pt] (A) at (0+ \startx,0+ \starty) {};

\node[ draw, circle, minimum size=20pt] (C) at (2*\scalex+ \startx,0+ \starty) {};

\node[draw, ellipse, minimum size=20pt, fill=red] (D) at (0.5*\scalex+ \startx,0.5*\scaley+ \starty) {};

\node[draw, ellipse, minimum size=20pt, fill=red] (E) at (1.5*\scalex+ \startx,0.5*\scaley+ \starty) {};

\node[ draw, ellipse, minimum size=20pt] (G) at (1*\scalex + \startx,1*\scaley+ \starty) {};

\draw [->,opacity=\opac] (A) -- (D);
\draw [->,opacity=\opac] (A) -- (E);
\draw [->,opacity=\opac] (C) -- (E);
\draw [->,opacity=\opac] (C) -- (D);
\draw [->,opacity=\opac] (D) -- (G);
\draw [->,opacity=\opac] (E) -- (G);
\end{tikzpicture}

}

%% file: graphs/neghyp_hyp.tex
\resizebox{50pt}{!}{
\begin{tikzpicture}[scale=0.6, every node/.style={scale=0.6}]
\def\scalex{2.1}
\def\scaley{2}
\def\startx{0}
\def\starty{-0.43}
\def\opac{1.0}
\def\starty{0}

\node[ draw, circle, minimum size=20pt] (A) at (0+ \startx,0+ \starty) {};

\node[ draw, circle, minimum size=20pt] (C) at (2*\scalex+ \startx,0+ \starty) {};

\node[draw, ellipse, minimum size=20pt, fill=red] (E) at (1.5*\scalex+ \startx,0.5*\scaley+ \starty) {};

\node[ draw, ellipse, minimum size=20pt] (G) at (1*\scalex + \startx,1*\scaley+ \starty) {};

\draw [->,opacity=\opac] (A) -- (G);
\draw [->,opacity=\opac] (A) -- (E);
\draw [->,opacity=\opac] (C) -- (E);
\draw [->,opacity=\opac] (E) -- (G);
\end{tikzpicture}

}

%% file: graphs/wordid_right.tex
\resizebox{50pt}{!}{
\begin{tikzpicture}[scale=0.6, every node/.style={scale=0.6}]
\def\scalex{2.1}
\def\scaley{2}
\def\startx{0}
\def\starty{-0.43}
\def\opac{1.0}
\def\starty{0}

\node[ draw, circle, minimum size=20pt] (A) at (0+ \startx,0+ \starty) {};

\node[ draw, circle, minimum size=20pt] (C) at (2*\scalex+ \startx,0+ \starty) {};

\node[draw, ellipse, minimum size=20pt, fill=red] (E) at (1.5*\scalex+ \startx,0.5*\scaley+ \starty) {};

\node[ draw, ellipse, minimum size=20pt] (G) at (1*\scalex + \startx,1*\scaley+ \starty) {};

\draw [->,opacity=\opac] (A) -- (G);
\draw [->,opacity=\opac] (C) -- (E);
\draw [->,opacity=\opac] (E) -- (G);
\end{tikzpicture}

}

%% file: graphs/subspace_wordid.tex
\resizebox{50pt}{!}{
\begin{tikzpicture}[scale=0.6, every node/.style={scale=0.6}]
\def\scalex{2.1}
\def\scaley{2}
\def\startx{0}
\def\starty{-0.43}
\def\opac{1.0}
\def\starty{0}

\node[ draw, circle, minimum size=20pt] (A) at (0+ \startx,0+ \starty) {};

\node[ draw, circle, minimum size=20pt] (C) at (2*\scalex+ \startx,0+ \starty) {};

\node[draw, ellipse, minimum size=20pt] (E) at (1.5*\scalex+ \startx,0.3*\scaley+ \starty) {};

\node[draw, ellipse, minimum size=20pt] (F) at (1.25*\scalex+ \startx,0.8*\scaley+ \starty) {};

\node[draw, ellipse, minimum size=20pt, fill=red] (G) at (1.75*\scalex+ \startx,0.8*\scaley+ \starty) {};

\node[draw, ellipse, minimum size=20pt] (H) at (1.5*\scalex+ \startx,1.3*\scaley+ \starty) {};

\node[ draw, ellipse, minimum size=20pt] (I) at (1*\scalex + \startx,1.8*\scaley+ \starty) {};

\draw [->,opacity=\opac] (A) -- (I);
\draw [->,opacity=\opac] (A) -- (E);
\draw [->,opacity=\opac] (C) -- (E);
\draw [->,opacity=\opac] (E) -- (F);
\draw [->,opacity=\opac] (E) -- (G);
\draw [->,opacity=\opac] (F) -- (H);
\draw [->,opacity=\opac] (G) -- (H);
\draw [->,opacity=\opac] (H) -- (I);
\end{tikzpicture}

}

%% file: tables/runtime.tex
\begin{table}[hp]
 \caption{Estimated runtime comparison between our method and brute force search (BFS) baseline (the number of testing hypotheses) for finding an alignment in a single targeted layer measured under the same settings. The runtime of DAS is invariant with the number of testing hypotheses.}
    \label{tab:runtime}
    \centering
    \begin{tabular}{lccc}
      \toprule
       & \multicolumn{3}{c}{\textbf{Runtime (sec)}} \\
      \textbf{Task} & \textbf{BFS} & \textbf{BFS}$_{\text{max}}$ & \textbf{DAS} \\
      \midrule
      Hierarchical Equality & 31 (32) & 6$e^{8}$ (${C^{32}_{16}}$) & 502 \\ [2ex]
      Monotonicity NLI & 198 (5) & 2$e^{58}$ (${C^{768}_{32}}$) & 1105 \\
      \bottomrule
    \end{tabular}
\end{table}

%% file: figures/RotationDegree.tex
\begin{figure}[ht]
\centering
\includegraphics[width=0.6\linewidth]{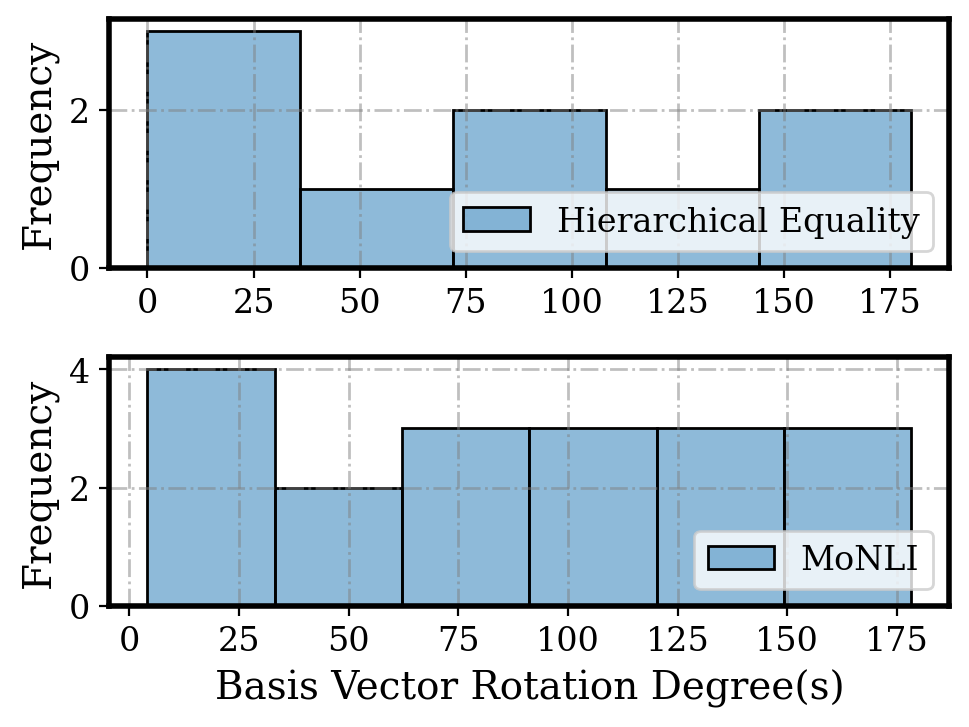}
\caption{Rotation measured in degree(s) of eigenvectors of the learned rotation matrix for each task.}
\label{fig:rotation-degree}
\end{figure}